\documentclass[twoside]{article}
\usepackage[accepted]{aistats2021}
\usepackage[round]{natbib}
\renewcommand{\bibname}{References}
\renewcommand{\bibsection}{\subsubsection*{\bibname}}
\usepackage{transparent,colortbl,xcolor}
\definecolor{red}{RGB}{255,0,81}
\definecolor{blue}{RGB}{0,139,251}
\usepackage{dsfont}
\usepackage{booktabs}
\usepackage{caption}
\usepackage{subfig}
\usepackage{hyperref}
\usepackage{url}
\usepackage{amsmath}
\usepackage{amssymb}
\usepackage{color}
\usepackage{amsthm}
\usepackage{newtxmath}
\usepackage{graphicx}
\usepackage{subfloat}
\usepackage{algorithm}
\usepackage{algpseudocode}
\usepackage[]{algorithmicx}
\newcommand*\Let[2]{\State #1 $\gets$ #2}
\algnewcommand{\LineComment}[1]{\State \(\triangleright\) #1}

\begin{document}

\runningtitle{}


\twocolumn[

\aistatstitle{Shapley Flow:\\ A Graph-based Approach to Interpreting Model Predictions}

\aistatsauthor{ Jiaxuan Wang \And Jenna Wiens \And  Scott Lundberg }

\aistatsaddress{ University of Michigan \\ \texttt{jiaxuan@umich.edu} \And  University of Michigan \\ \texttt{wiensj@umich.edu} \And Microsoft Research \\ \texttt{scott.lundberg@microsoft.com}} ]

% \author{Jiaxuan Wang, Jenna Wiens \\ %\thanks{ Use footnote for providing further information
% %about author (webpage, alternative address)---\emph{not} for acknowledging
% %funding agencies.  Funding acknowledgements go at the end of the paper.} \\
% Department of Computer Science and Engineering\\
% University of Michigan\\
% 2260 Hayward St, Ann Arbor, MI 48109, USA \\
% \texttt{\{jiaxuan,wiensj\}@umich.edu} \\
% \And
% Scott Lundberg \\
% Microsoft Research \\
% %University of the Witwatersrand \\
% 14820 NE 36th St, Redmond, WA 98052, USA \\
% \texttt{Scott.Lundberg@microsoft.com}
% }

\begin{abstract}
Many existing approaches for estimating feature importance are problematic because they ignore or hide dependencies among features. A causal graph, which encodes the relationships among input variables, can aid in assigning feature importance. However, current approaches that assign credit to nodes in the causal graph fail to explain the entire graph. In light of these limitations, we propose Shapley Flow, a novel approach to interpreting machine learning models. It considers the entire causal graph, and assigns credit to \textit{edges} instead of treating nodes as the fundamental unit of credit assignment. Shapley Flow is the unique solution to a generalization of the Shapley value axioms for directed acyclic graphs. We demonstrate the benefit of using Shapley Flow to reason about the impact of a model's input on its output. In addition to maintaining insights from existing approaches, Shapley Flow extends the flat, set-based, view prevalent in game theory based explanation methods to a deeper, \textit{graph-based}, view. This graph-based view enables users to understand the flow of importance through a system, and reason about potential interventions.
%\vspace{-10pt}
%when causal links can be broken. %[need the so what sentence - big picture what does this mean to other ML researchers]

%While traditional feature attribution methods aim at attributing changes in output to changes in input, we note that the structure within input variables is crucial for interpretation as well.
%Understanding a model from existing feature attribution methods is insufficient because they either hide the impact of upstream causes or downstream dependencies of features, which prevent us from reasoning about the robustness of the system under intervention. Therefore, we propose Shapley Flow to assign credit along the edges of a causal graph instead of on the features. We demonstrate the benefit of Shapley Flow on two real datasets. Furthermore, our proposed approach is unique under an extension of Shapley value's axioms for graphs. Our approach unifies three Shapley value based model interpretation methods already in use. 
\end{abstract}

\section{Introduction}

\begin{figure}
\centering
\includegraphics[width=0.5\linewidth]{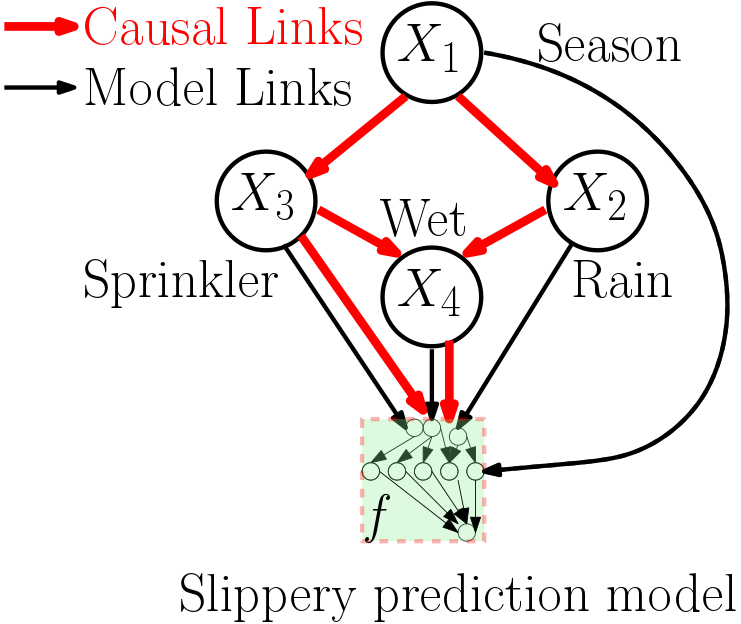}
\setlength{\belowcaptionskip}{-14pt}
\caption{Causal graph for the sprinkler example from Chapter 1.2 of \cite{pearl2009causality}. The model, $f$, can be expanded into its own graph. To simplify the exposition, although $f$ takes $4$ variables as input, we arbitrarily assumed that it only depends on $X_3$ and $X_4$ directly (\textit{i.e.}, $f(X_1, X_2, X_3, X_4)=g(X_3, X_4)$ for some $g$).}
\label{fig:pearl_causal_graph}
\end{figure}

Explaining a model's predictions by assigning importance to its inputs (\textit{i.e.}, feature attribution) is critical to many applications in which a user interacts with a model to either make decisions or gain a better understanding of a system \citep{simonyan2013deep, lundberg2017unified,zhou2016learning,shrikumar2017learning, baehrens2010explain, binder2016layer, springenberg2014striving, sundararajan2017axiomatic, fisher2018all, breiman2001random}. However, correlation among input features presents a challenge when estimating feature importance. 

Consider a motivating example adapted from \cite{pearl2009causality}, in which we are given a model $f$ that takes as input four features: the season of the year ($X_1$), whether or not it's raining ($X_2$), whether the sprinkler is on ($X_3$), and whether the pavement is wet ($X_4$) and outputs a prediction $f(\mathbf{x})$, representing the probability that the pavement is slippery (capital $X$ denotes a random variable; lower case $\mathbf{x}$ denotes a particular sample). Assume, the inputs are related through the causal graph in \textbf{Figure \ref{fig:pearl_causal_graph}}. When assigning feature importance, existing approaches that ignore this causal structure \citep{janzing2020feature, sundararajan2019many, datta2016algorithmic} assign zero importance to the season, since it only indirectly affects the outcome through the other input variables. However, such a conclusion may lead a user astray - since changing $X_1$ would most definitely affect the outcome. 

Recognizing this limitation, researchers have recently proposed approaches that leverage the causal structure among the input variables when assigning credit \citep{frye2019asymmetric,heskes2020causal}. However, such approaches provide an incomplete picture of a system as they only assign credit to nodes in a graph. For example, the ASV method of \citet{frye2019asymmetric} solves the earlier problem of ignoring indirect or upstream effects, but it does so by ignoring direct or downstream effects. In our example, season would get all the credit despite the importance of the other variables. This again may lead a user astray - since intervening on $X_3$ or $X_4$ would affect the outcome, yet they are given no credit. The Causal Shapley values of \citet{heskes2020causal} do assign credit to $X_3$ and $X_4$, but force this credit to be divided with $X_1$. This leads to the problem of features being given less importance simply because their downstream variables are also included in the graph.
%s problematic because users of ASV wouldn't be able tell between a model that has learned useful concepts and a model that just picks up on features that are correlated to them.

\begin{figure}%[H]
\centering
%\subfloat{\label{fig:boundary_independent}
\includegraphics[width=0.32\linewidth]{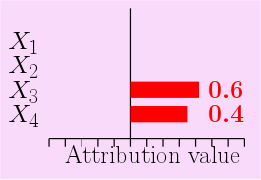}
%}
%\subfloat{\label{fig:boundary_asv}
\includegraphics[width=0.32\linewidth]{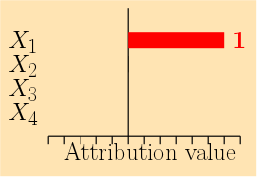}
%}
%\subfloat{\label{fig:boundary_flow}
\includegraphics[width=0.32\linewidth]{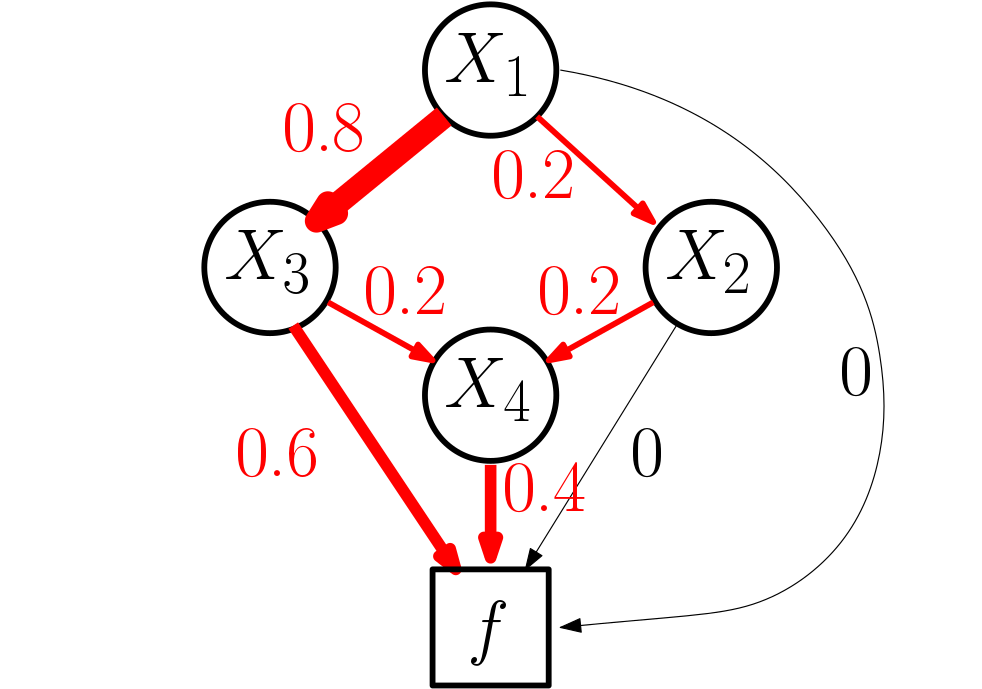}%}

\subfloat[Independent]{\label{fig:illustration_independent}
\includegraphics[width=0.32\linewidth]{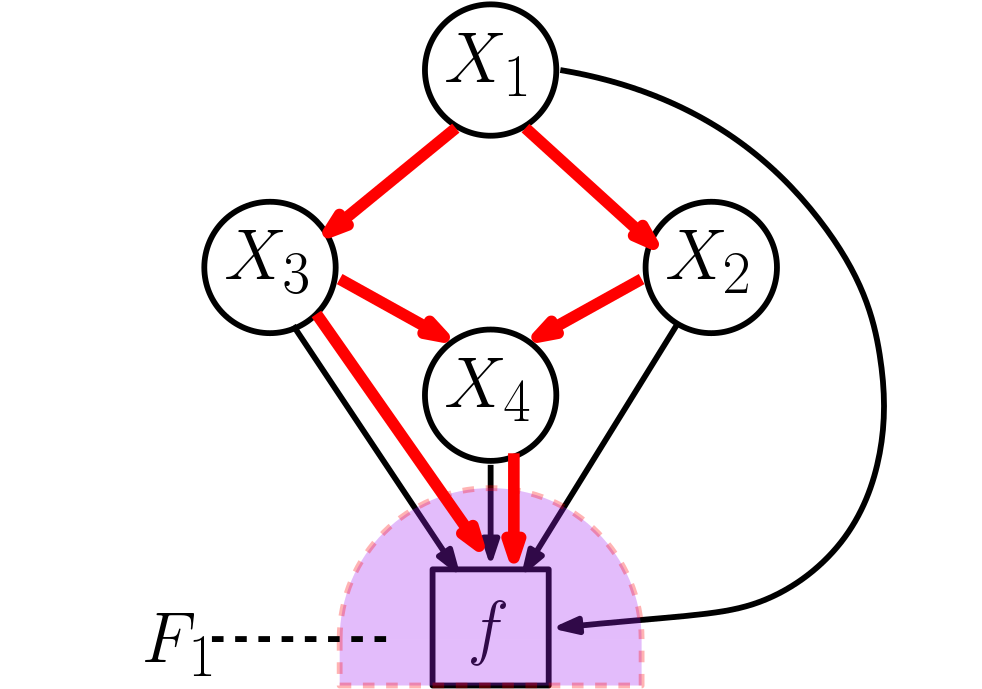}
}
\subfloat[ASV]{\label{fig:illustration_asv}
\includegraphics[width=0.32\linewidth]{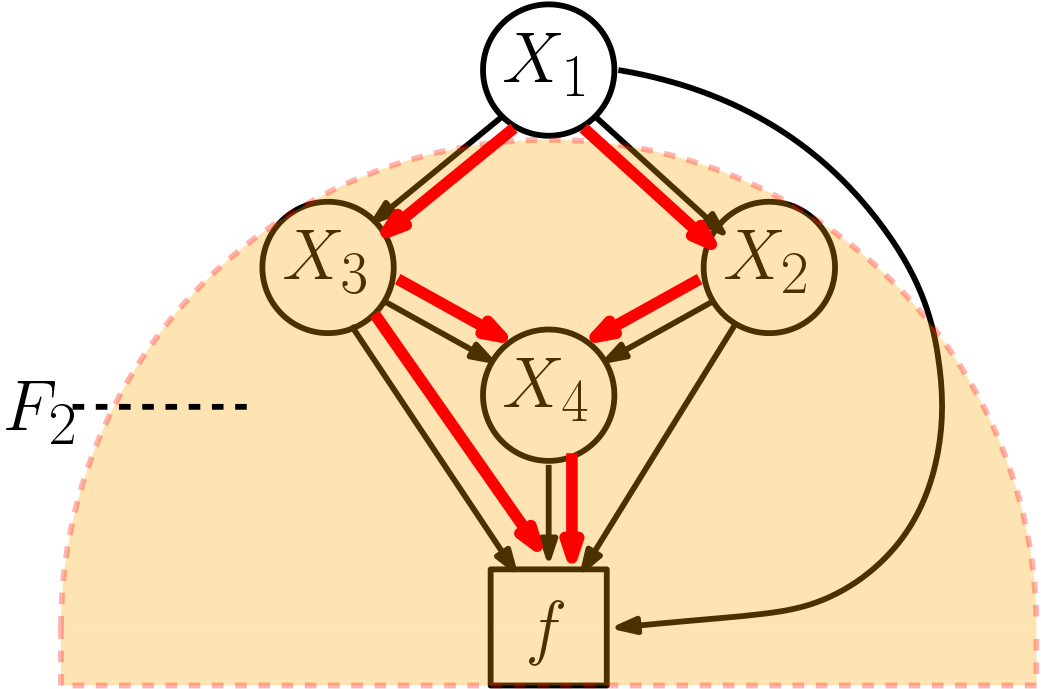}
}
\subfloat[Shapley Flow]{\label{fig:illustration_flow}
\includegraphics[width=0.32\linewidth]{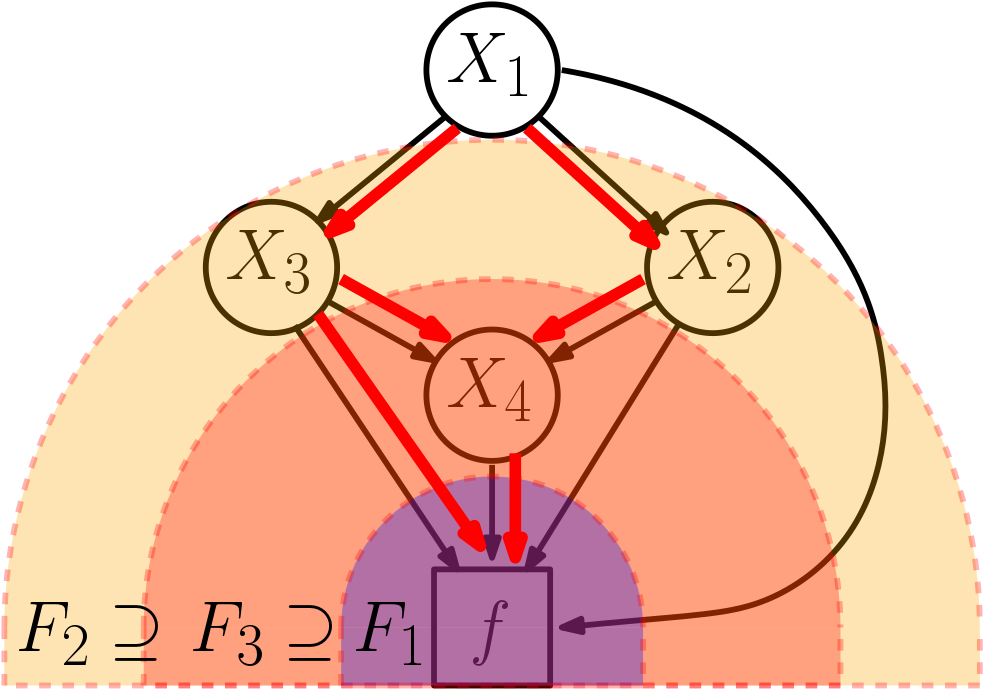}
}

\setlength{\belowcaptionskip}{-14pt}
\caption{
Top: Output of attribution methods for the example in \textbf{Figure \ref{fig:pearl_causal_graph}}. Bottom: Causal structure (black edges) and explanation boundaries used by each method. As a reference, we copied the true causal links (red) from \textbf{Figure \ref{fig:pearl_causal_graph}}. An explanation boundary $\mathcal{B}:=(D, F)$ is a cut in the graph that defines a ``model'' $F$ (nodes in the shaded area in each figure) to be explained. Refer to \textbf{Section \ref{sec:feature_causal}} for a detailed discussion.}
\label{fig:pearl_example}
\end{figure}

Given that current approaches end up ignoring or dividing either downstream (\textit{i.e.}, direct) or upstream (\textit{i.e.}, indirect) effects, we develop Shapley Flow, a comprehensive approach to interpreting a model (or system) that incorporates the causal relationship among input variables, while accounting for both direct and indirect effects. In contrast to prior work, we accomplish this by reformulating the problem as one related to assigning credit to \textit{edges} in a causal graph, instead of \textit{nodes} (\textbf{Figure \ref{fig:illustration_flow}}). Our key contributions are as follows.

\begin{itemize}
    \item We propose the first (to the best of our knowledge) generalization of Shapley value feature attribution to graphs, providing a complete system-level view of a model.%'s reasoning.  
    \vspace{-7pt}
    \item Our approach unifies three previous game theoretic approaches to estimating feature importance.
    \vspace{-7pt}
    \item Through examples on real data, we demonstrate how our approach facilitates understanding feature importance.
\end{itemize}

In this work, we take an axiomatic approach motivated by cooperative game theory, extending Shapley values to graphs. The resulting algorithm, Shapley Flow, generalizes past work on estimating feature importance \citep{lundberg2017unified,frye2019asymmetric,lopez2009relationship}. The estimates produced by Shapley Flow represent the unique allocation of credit that conforms to several natural axioms. Applied to real-world systems, Shapley Flow can help a user understand both the direct and indirect impact of changing a variable, generating insights beyond current feature attribution methods.

\section{Problem Setup \& Background}

Given a model, or more generally a system, that takes a set of inputs and produces an output, we focus on the problem of quantifying the effect of each input on the output. Here, building off previous work, we formalize the problem setting.

\subsection{Problem Setup} \label{sec:feature_attribution}

Quantifying the effect of each input on a model's output can be formulated as a credit assignment problem. Formally, given a target sample input $\boldsymbol{x}$, a background sample input $\boldsymbol{x'}$, and a model $f: \mathbb{R}^d \rightarrow \mathbb{R}$, we aim to explain the difference in output \textit{i.e.}, $f(\boldsymbol{x}) - f(\boldsymbol{x'})$. We assume $\boldsymbol{x}$ and $\boldsymbol{x}'$ are of the same dimension $d$, and each entry can be either discrete or continuous.

We also assume access to a causal graph, as formally defined in Chapter 6 of \cite{peters2017elements}, over the $d$ input variables. Given this graph, we seek an assignment function $\phi$ that assigns credit $\phi(e) \in \mathbb{R}$ to each edge $e$ in the causal graph such that they collectively explain the difference $f(\boldsymbol{x}) - f(\boldsymbol{x'})$. In contrast with the classical setting \citep{lundberg2017unified,sundararajan2017axiomatic,frye2020shapley,aas2019explaining} in which credit is placed on features (\textit{\i.e.}, seeking a node assignment function $\psi(i) \in \mathbb{R}$ for $i \in [1\cdots d]$), our edge-based approach is more flexible because we can recover node $i$'s importance by defining $\psi(i) = \sum_{e \in \text{i's outgoing edges}} \phi(e)$. This exactly matches the classic Shapley axioms \citep{shapley1953value} when the causal graph is degenerate with a single source node connected directly to all the input features.

Here, the effect of the input on the output is measured with respect to a background sample. For example, in a healthcare setting, we may set the features in the background sample to values that are deemed typical for a disease. We assume a single background value for notational convenience, but the formalism easily extends to the common scenario of multiple background values or a distribution of background values, $P$, by defining the explanation target to be $f(\boldsymbol{x}) - \mathbb{E}_{\boldsymbol{x'} \sim P} f(\boldsymbol{x'})$.

\subsection{Feature Attribution with a Causal Graph} \label{sec:feature_causal}

In our problem setup, we assume access to a causal graph, which can help in reasoning about the relationship among input variable. However, even with a causal graph, feature attribution remains challenging because it is unclear how to rightfully allocate credit for a prediction among the nodes and/or edges of the graph. Marrying interpretation with causality is an active field (see \cite{moraffah2020causal} for a survey). A causal graph in and of itself does not solve feature attribution. While a causal graph can be used to answer a specific question with a specific counterfactual, summarizing many counterfactuals to give a comprehensive picture of the model is nontrivial. Furthermore, each node in a causal graph could be a blackbox model that needs to be explained. To address this challenge, we generalize game theoretic fairness principles to graphs.% through what we call the \textbf{boundary of explanation}.% a unifying perspective based on the boundary of explanation.

Given a graph, $\mathcal{G}$, that consists of a causal graph over the  the model of interest $f$ and its inputs, we define the \textbf{boundary of explanation} as a cut $\mathcal{B}:=(D, F)$ that partitions the input variables and the output of the model (\textit{i.e.}, the nodes of the graph) into  $D$ and $F$ where source nodes (nodes with no incoming edges) are in $D$ and sink nodes (nodes with no outgoing edges) are in $F$. Note that $\mathcal{G}$ has a single sink, $f(\boldsymbol{x}) \in \mathbb{R}$. A cut set is the set of edges with one endpoint in $D$ and another endpoint in $F$, denoted as $cut(\mathcal{B})$. It is helpful to think of $F$ as an alternative model definition, where a boundary of explanation (\textit{i.e.}, a model boundary) defines what part of the graph we consider to be the ``model''. If we collapse $F$ into a single node that subsumes $f$, then $cut(\mathcal{B})$ represents the direct inputs to this new model.

Depending on the causal graph, multiple boundaries of explanation may exist. Recognizing this multiplicity of choices helps shed light on an ongoing debate in the community regarding feature attribution and whether one should perturb features while staying on the data manifold or perturb them independently \citep{chen2020true,janzing2020feature,sundararajan2019many}. On one side, many argue that perturbing features independently reveals the functional dependence of the model, and is thus \textit{true to the model} \citep{janzing2020feature, sundararajan2019many, datta2016algorithmic}. However, independent perturbation of the data can create unrealistic or invalid sets of model input values. Thus, on the other side, researchers argue that one should perturb features while staying on the data manifold, and so be \textit{true to the data} \citep{aas2019explaining, frye2019asymmetric}. However, this can result in situations in which features not used by the model are given non-zero attribution. Explanation boundaries help us unify these two viewpoints. As illustrated in \textbf{Figure \ref{fig:illustration_independent}}, when we independently perturb features, we assume the causal graph is flat and the explanation boundary lies between $\boldsymbol{x}$ and $f$ (\textit{i.e.}, $D$ contains all of the input variables). In this example, since features are assumed independent all credit is assigned to the features that directly impact the model output, and indirect effects are ignored (no credit is assigned to $X_1$ and $X_2$). In contrast, when we perform on-manifold perturbations with a causal structure, as is the case in Asymmetric Shapley Values (ASV) \citep{frye2019asymmetric}, all the credit is assigned to the source node because the source node determines the value of all nodes in the graph (\textbf{Figure \ref{fig:illustration_asv}}). This results in a different boundary of explanation, one between the source nodes and the remainder of the graph. Although giving $X_1$ credit does not reflect the true functional dependence of $f$, it does for the model defined by $F_2$ (\textbf{Figure \ref{fig:illustration_flow}}). Perturbations that were previously faithful to the data are faithful to a  ``model'', just one that corresponds to a different boundary. See \textbf{Section \ref{sec:on_manifold_boundary}} in the Appendix for how on-manifold perturbation (without a causal graph) can be unified using explanation boundaries.

%observing any feature could affect all of its correlated inputs to the data (nodes with a tilde), effectively forming a 2-level graph shown in . Now, although $\tilde{X_1}$ has no effect on $f$, $X_1$ can impact the output through its correlation with other input nodes. If we define let $D$ include all source nodes, we are again true to the modeled defined by $F$. %additional source nodes (e.g., $U$ in \textbf{Figure \ref{fig:illustration_on_manifold}}) capture the correlations among input variables, and thus $D$ ends up consisting only of those source nodes. 
%The credit is split equally between indirect effects and direct effects, even though there is no reason to reduce the credit for a direct effect if we happen to also observe it's cause. 
% Other methods can be unified by explanation boundary as well. For example, 

Beyond the boundary directly adjacent to the model of interest, $f$, and the boundary directly adjacent to the source nodes, there are other potential boundaries (\textbf{Figure \ref{fig:illustration_flow}}) a user may want to consider. However, simply generating explanations for each possible boundary can quickly overwhelm the user (\textbf{Figures \ref{fig:illustration_independent},  \ref{fig:illustration_asv}} in the main text, and \textbf{\ref{fig:illustration_on_manifold}} in the Appendix). Our approach sidesteps the issue of selecting a single explanation boundary by considering all explanation boundaries simultaneously. This is made possible by assigning credit to the edges in a causal graph (\textbf{Figure \ref{fig:illustration_flow}}). Edge attribution is strictly more powerful than feature attribution because we can simultaneously capture the direct and indirect impact of edges. We note that concurrent work by \cite{heskes2020causal} also recognized that existing methods have difficulty capturing the direct and indirect effects simultaneously. Their solution however is node based, so it is forced to split credit between parents and children in the graph.% still node based, thus not able to solve the problem. % In addition to covering existing boundaries, edge attribution also considers boundaries not yet explored by current methods, giving users a complete picture of the system. %We can recover feature attribution by summing attributions of the outgoing edges of a node. %Note that under a flat graph, node attribution and edge attribution are exactly the same because they are associated in a one to one mapping.

While other approaches to assign credit on a graph exist, (\textit{e.g.}, Conductance from \cite{dhamdhere2018important} and DeepLift from \cite{shrikumar2016not}), they were proposed in the context of understanding internal nodes of a neural network, and depend on implicit linearity and continuity assumptions about the model. We aim to understand the causal structure among the input nodes in a fully model agnostic manner, where  discrete variables are allowed, and no differentiability assumption is made. To do this we generalize the widely used Shapley value \citep{adadi2018peeking,mittelstadt2019explaining,lundberg2018consistent,sundararajan2019many,frye2019asymmetric,janzing2020feature,chen2020true} to graphs.

\section{Proposed Approach: Shapley Flow} \label{sec:method}

Our proposed approach, Shapley Flow, attributes credit to edges of the causal graph. In this section, we present  the intuition behind our approach and then formally show that it uniquely satisfies a generalization of the classic Shapley value axioms, while unifying previously proposed approaches.

\subsection{Assigning Credit to Edges: Intuition} \label{sec:method_intuition}

\begin{figure}%[H]
\centering
\subfloat[$e_2$ updates after $e_1$]{\label{fig:edge_importance}
\includegraphics[width=0.95\linewidth]{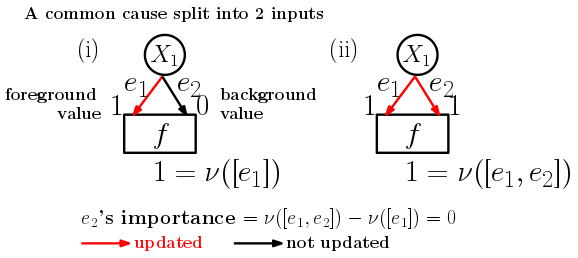}
}

\subfloat[$e_2$ updates before $e_1$]{\label{fig:or_func}
\includegraphics[width=0.8\linewidth]{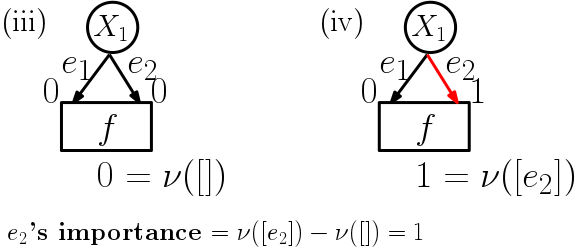}
}
\setlength{\belowcaptionskip}{-14pt}
\caption{Edge importance is measured by the change in output when an edge is added. When a model is non-linear, say $f=OR$, we need to average over all scenarios in which $e_2$ can be added to gauge its importance. \textbf{Section \ref{sec:method_intuition}} has a detailed discussion.} % \textbf{(a)} A natural approach is to compare the difference in output with and without the edge (\textit{i.e.}, (i) vs. (ii)). We model the edge as a channel carrying the value of its source node. Removing an edge prevents messages from being received from the edge, and so forces the edge's target node to use the source's background value to compute its output. \textbf{(b)} Message ordering matters. Consider $f=\text{OR}$ function with $\boldsymbol{x}=\boldsymbol{1}$ and $\boldsymbol{x'}=\boldsymbol{0}$ as input. If we always calculate the difference in output upon receiving message through $e_2$ after $e_1$, (\textit{i.e.}, from (i) to (ii) output changes by $0$), we would give $e_2$ $0$ importance, despite that it is important when its message is received first (\textit{i.e.}, from (iii) to (iv) output changes by $1$). Therefore we need to average over all scenarios in which an edge is added.}
\label{fig:or_func2}
\end{figure}

Given a causal graph defining the relationship among input variables, we re-frame the problem of feature attribution to focus on the edges of a graph rather than nodes. Our approach results in edge credit assignments as shown in \textbf{Figure \ref{fig:illustration_flow}}. As mentioned above, this eliminates the need for multiple explanations (\textit{i.e.}, bar charts) pertaining to each explanation boundary. Moreover, it allows a user to better understand the nuances of a system by providing information regarding what would happen if a single causal link breaks.

{\bf Shapley Flow is the unique assignment of credit to edges such that a relaxation of the classic Shapley value axioms are satisfied for all possible boundaries of explanation.} Specifically, we extend the efficiency, dummy, and linearity axioms from \cite{shapley1953value} and add a new axiom related to boundary consistency. Efficiency states that the attribution of edges on any boundary must add up to $f(\mathbf{x}) - f(\mathbf{x'})$. Linearity states that explaining a linear combination of models is the same as explaining each model, and linearly combining the resulting attributions. Dummy states that if adding an edge does not change the output in any scenarios, the edge should be assigned $0$ credit. Boundary consistency states that edges shared by different boundaries need to have the same attribution when explained using either boundary. These concepts are illustrated in \textbf{Figure \ref{fig:axioms}} and formalized in \textbf{Section \ref{sec:axioms}}.

%Besides an axiomatic construction, Shapley Flow can also be intuitively derived. 
An edge is important if removing it causes a large change in the model's prediction. However, what does it mean to remove an edge? If we imagine every edge in the graph as a channel that sends its source node's current value to its target node, then removing an edge $e$ simply means messages sent through $e$ fail. In the context of feature attribution, in which we aim to measure the difference between $f(\mathbf{x})-f(\mathbf{x'})$, this means that $e$'s target node still relies on the source's background value in $\mathbf{x'}$ to update its current value, as opposed to the source node's foreground value in $\mathbf{x}$, as illustrated in \textbf{Figure \ref{fig:edge_importance}}. Note that treating edge removal as replacing the parent node with the background value is equivalent to the approach advocated by \cite{janzing2020feature}, and matches the default behavior of SHAP and related methods. However, we cannot simply toggle edges one at a time. Consider a simple OR function $g(X_1, X_2) = X_1 \vee X_2$, with $x_1=1$, $x_2=1$, $x_1'=0$, $x_2'=0$. Removing either of the edges alone, would not affect the output and both  $x_1$ and $x_2$ would be (erroneously) assigned $0$ credit.

To account for this, we consider all scenarios (or partial histories) in which the edge we care about can be added (see \textbf{Figure \ref{fig:or_func}}). Here, $\nu$ is a function that takes a list of edges and evaluates the network with edges updated in the order specified by the list. For example, $\nu([e_1])$ corresponds to the evaluation of $f$ when only $e_1$ is updated. Similarly $\nu([e_1, e_2])$ is the evaluation of $f$ when $e_1$ is updated followed by $e_2$. The list $[e_1, e_2]$ is also referred to as a (complete) \textit{history} as it specifies how $\mathbf{x'}$ changes to $\mathbf{x}$.
 %To account for this, we assume the input variables are updated one at a time at discrete time steps, where multiple update orderings (or execution histories) are possible 

For the same edge, attributions derived from different explanation boundaries should agree, otherwise simply including more details of a model in the causal graph would change upstream credit allocation, even though the model implementation was unchanged. We refer to this property as \textit{boundary consistency}. The Shapley Flow value for an edge is the difference in model output when removing the edge averaged over all histories that are boundary consistent (as defined below).

\subsection{Model explanation as value assignments in games} \label{sec:game}

The concept of Shapley value stems from game theory, and has been extensively applied in model interpretability \citep{vstrumbelj2014explaining,datta2016algorithmic,lundberg2017unified,frye2019asymmetric,janzing2020feature}. Before we formally extend it to the context of graphs, we define the credit assignment problem from a game theoretic perspective. 

%Here, we model the flow of information $\mathcal{G}$ as a distributed network. Each node and edge in the network corresponds to a node and edge in $\mathcal{G}$. Every node can receive and emit information through directed edges but each edge could have arbitrary delay (thus leading to multiple possible histories). Upon receiving a new message (output of the edge's source node), the edge's target node uses its most recently received messages from each incoming edge, starting from $\mathbf{x'}$ of the edge's source node, to compute a new message and broadcast it out. The messages sent through the same edge are received in the same order.
%Given this model, 

Given the message passing system in \textbf{Section \ref{sec:method_intuition}}, we formulate the credit assignment problem as a game specific to an explanation boundary $\mathcal{B}:=(D, F)$. The game consists of a set of players $\mathcal{P}_{\mathcal{B}}$, and a payoff function $\nu_{\mathcal{B}}$. We model each edge external to $F$ as a player.
%so that computation in $F$ can be arbitrarily implemented. 
A \textit{history} is a list of edges detailing the event from $t=0$ (values being $\mathbf{x'}$)  to $t=T$ (values being $\mathbf{x}$). For example, the history $[i, j, i]$ means that the edge $i$ finishes transmitting a message containing its source node's most recent value to its target node, followed by the edge $j$, and followed by the edge $i$ again. A \textit{coalition} is a partial history from $t=0$ to any $t \in [0 \cdots T]$. The \textit{payoff function}, $\nu$, associates each coalition with a real number, and is defined in our case as the evaluation of $F$ following the coalition. %Note that any message sent through edges not in the cut set defined by $\mathcal{B}$ have no affect on the model's output because the input to $F$ has not changed. To assign credit to those edges, we must consider explanation boundaries for which they are included in the cut set. 

This setup is a generalization of a typical cooperative game in which the ordering of players does not matter (only the set of players matters). However, given our message passing system, history is important. % An example is "Y" shaped causal graph (e.g. a->b, c->b, b->d), the v([a->b, c->b, b->d]) is different from v([c->b, a->b, b->d]) because b->d would update a->b's message in the first case, while updating b->d's message in the latter case.
 In the following sections, we denote `$+$' as list concatenation,  `$[]$' as an empty coalition, and $\mathcal{H}_{\mathcal{B}}$ as the set of all possible histories. We denote $\tilde{\mathcal{H}}_{\mathcal{B}} \subseteq \mathcal{H}_{\mathcal{B}}$ as the set of boundary consistent histories. The corresponding coalitions for $\mathcal{H}_{\mathcal{B}}$ and $\tilde{\mathcal{H}}_{\mathcal{B}}$ are denoted as $\mathcal{C}_{\mathcal{B}}$ and 
$\tilde{\mathcal{C}}_{\mathcal{B}}$ respectively. A sample game setup is illustrated in \textbf{Figure \ref{fig:or_func2}}.

\subsection{Axioms} \label{sec:axioms}

\begin{figure}%[H]
\centering
\subfloat[Effi. $+$ Bound. Consist.]{\label{fig:efficiency_and_boundary_consistency}
\includegraphics[width=0.45\linewidth]{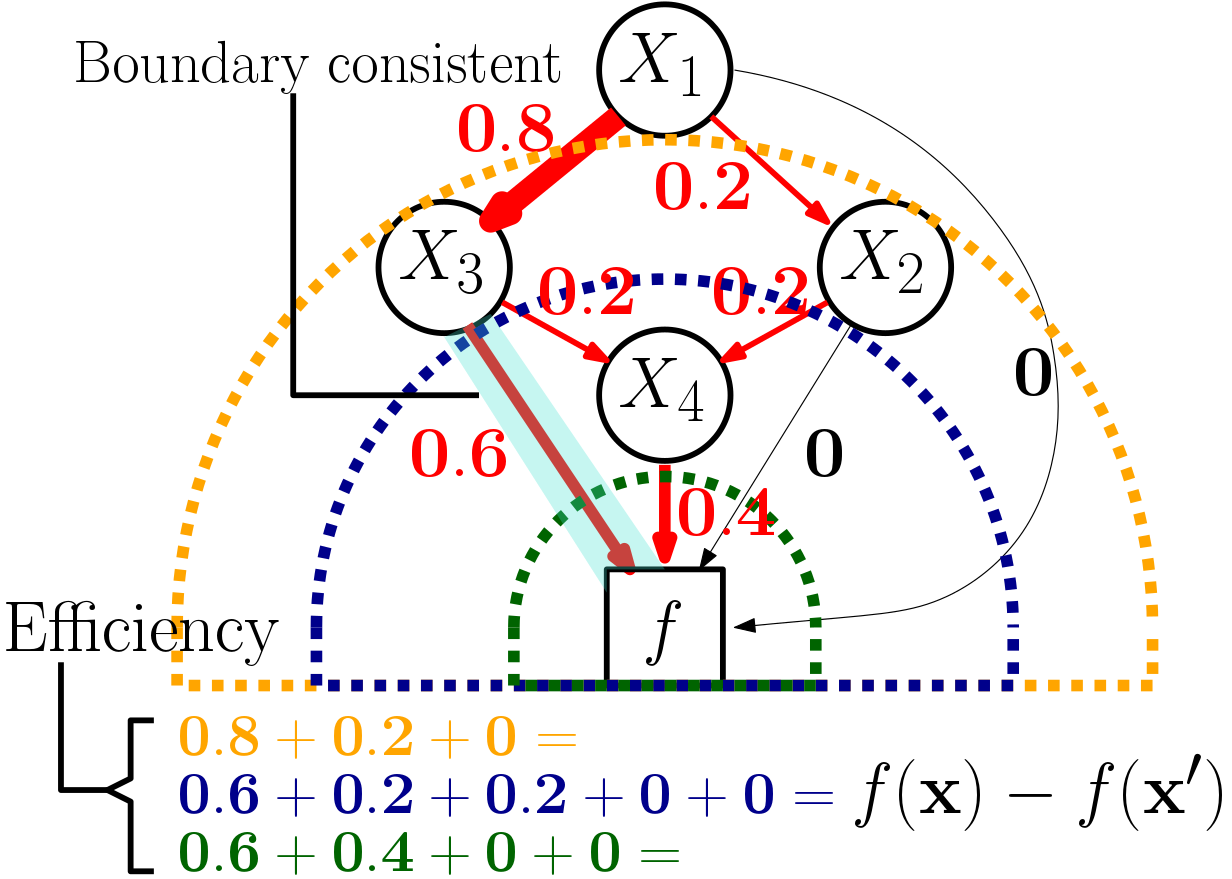}}
\subfloat[Dummy player]{\label{fig:dummy}
\includegraphics[width=0.5\linewidth]{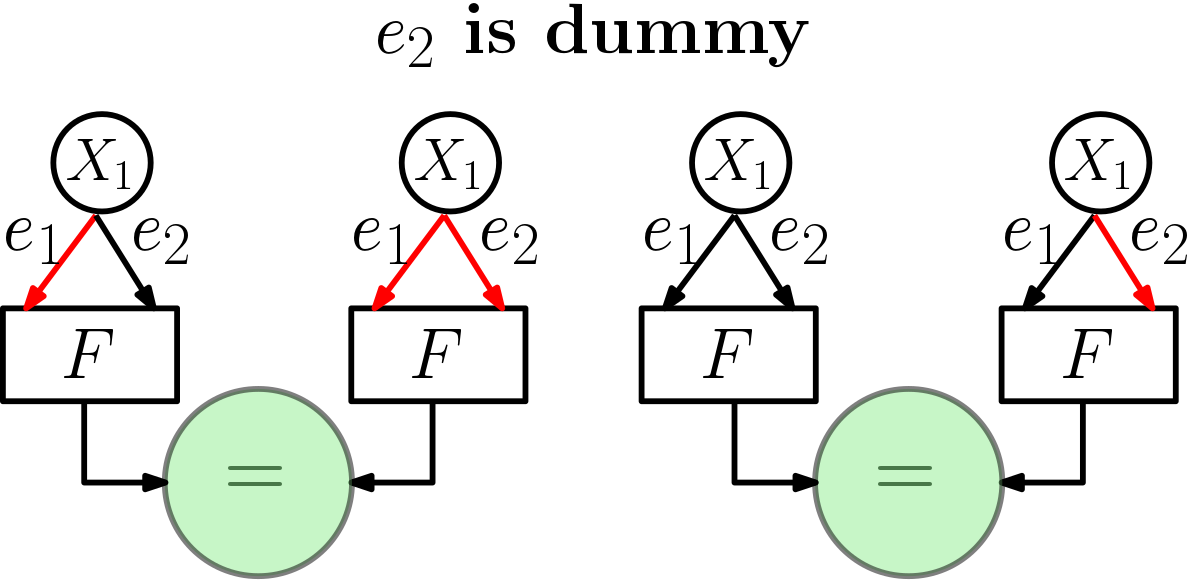}}

\subfloat[Linearity]{\label{fig:linearity}
\includegraphics[width=0.95\linewidth]{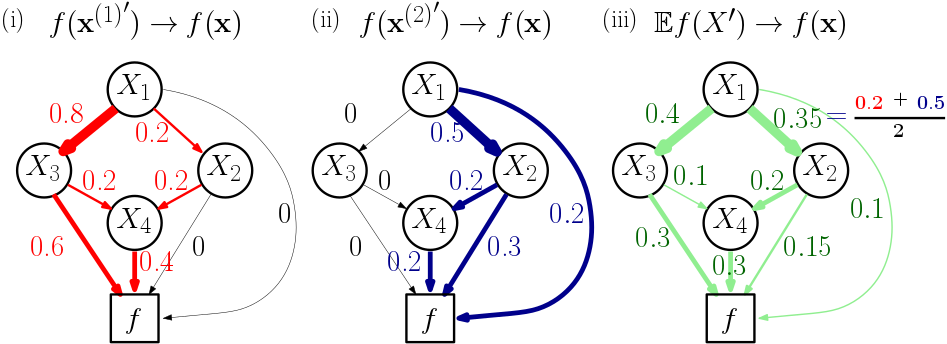}}
\setlength{\belowcaptionskip}{-10pt}

\caption{Illustration for axioms for Shapley Flow. Except for boundary consistency, all axioms stem from Shapley value's axioms \citep{shapley1953value}. Detailed explanations are included in \textbf{Section \ref{sec:axioms}}.} %\textbf{(a)} \textit{Efficiency}: attributions for any boundary should add up to $f(\mathbf{x}) - f(\mathbf{x'})$. This implies that the sum of attributions in equals to the sum of attributions out for every node. \textit{Boundary consistency}: observe that the edge wrapped by a teal band is shared by both the blue and green boundaries, forcing them to give the same attribution to the edge. \textbf{(b)} \textit{Dummy player}: message passing through $X_2$'s edge does not change the output in all histories, thus it should be given $0$ attribution. \textbf{(c)} \textit{Linearity}: the attribution associated with a payoff function (iii) defined by a linear combination of 2 other payoff functions (i) and (ii) is the the same linear combination of (i) and (ii)'s attributions. One can use linearity to explain $f(\mathbf{x}) - \mathbb{E}(f(X'))$ by independent computing attributions for each background sample $\mathbf{x^{(i)'}}$ and then take the average of the attributions, without recomputing from scratch whenever the background sample's distribution changes.}
\label{fig:axioms}
\end{figure}

%Given this game theoretic perspective, w
We formally extend the classic Shapley value axioms (efficiency, linearity, and dummy) and include one additional axiom, the boundary consistency axiom, that connects all boundaries together.

\begin{itemize}
    \item Boundary consistency: for any two boundaries $\mathcal{B}_1=(D_1, F_1)$ and $\mathcal{B}_2=(D_2, F_2)$, $\phi_{\nu_{\mathcal{B}_1}}(i) = \phi_{\nu_{\mathcal{B}_2}}(i)$ for $i \in cut(\mathcal{B}_1) \cap cut(\mathcal{B}_2)$
    
    For edges that are shared between boundaries, their attributions must agree. In \textbf{Figure \ref{fig:efficiency_and_boundary_consistency}}, the edge wrapped by a teal band is shared by both the blue and green boundaries, forcing them to give the same attribution to the edge.
\end{itemize}

\begin{figure}%[H]
\centering
\includegraphics[width=0.98\linewidth]{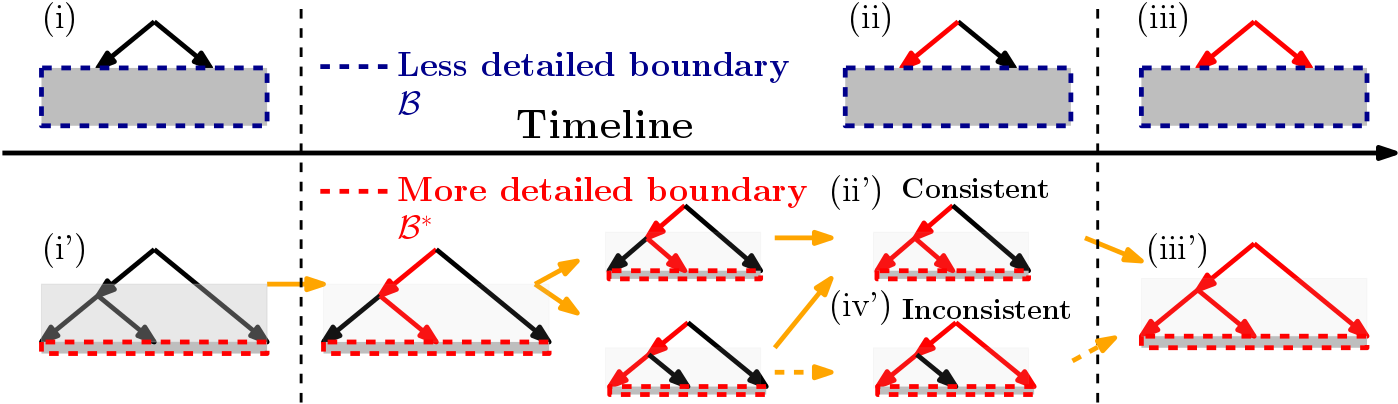}

\setlength{\belowcaptionskip}{-10pt}
\caption{Boundary Consistency. For the blue boundary (upper), we show one potential history $h$. When we expand $h$ to the red boundary (lower), $h$ corresponds to multiple histories as long as each history contains states that match (i) (ii) and (iii). (i') matches (i), no messages are received in both states. (ii') matches (ii), the full impact of message transmitted through the left edge is received at the end of computation. (iii') matches (iii), all messages are received. In contrast, the history containing (iv') has no state matching (ii), and thus is inconsistent with $h$.} % On the top, we show a specific history, associated with a boundary represented by the black box. The message sent along the left (right) edge is colored blue (brown). In the bottom, we open up the box and show the expanded set of histories. The vertical bars show how the time steps are aligned. When the blue message enters the box, the black box computes its output immediately before the brown message arrives (that's how the payoff function for the boundary is defined). That is the computation through the right edge can only begin after the computation through the left edge completes. Therefore, in the expanded histories, we cannot reach the state highlighted in brown because the impact of the blue message only partially reaches the output before the arrival of the brown message.}
\label{fig:boundary_consistency}
\end{figure}

In the general setting, not all credit assignments are boundary consistent; different boundaries could result in different attributions for the same edge\footnote{We include an example in the Appendix \textbf{Section \ref{sec:inconsistent_history}} to demonstrate why considering all histories $\mathcal{H}$ can violate boundary consistency, thus motivating the need to only focus on boundary consistent histories.}. This occurs when histories associated with different boundaries are inconsistent  (\textbf{Figure \ref{fig:boundary_consistency}}). Moving the boundary from $\mathcal{B}$ to $\mathcal{B}^*$ (where $\mathcal{B}^*$ is the boundary with $D$ containing $f$'s inputs), results in a more detailed set of histories. This expansion has $2$ constraints. First, any history in the expanded set follows the message passing system in \textbf{Section \ref{sec:method_intuition}}. Second, when a message passes through the boundary, it  immediately reaches the end of computation as $F$ is assumed to be a black-box.

Denoting the history expansion function into $\mathcal{B}^*$ as $HE$ (\textit{i.e.}, $HE$ takes a history $h$ as input and expand it into a set of histories in $\mathcal{B}^*$ as output) and denoting the set of all boundaries as $\mathcal{M}$, a history $h$ is \textit{boundary consistent} if $\exists h_{\mathcal{B}} \in \mathcal{H}_{\mathcal{B}}$ for all $\mathcal{B} \in \mathcal{M}$ such that 
$$(\bigcap_{\mathcal{B} \in \mathcal{M}} HE(h_{\mathcal{B}})) \cap HE(h) \neq \emptyset$$
That is $h$ needs to have at least one fully detailed history in which all boundaries can agree on. $\tilde{\mathcal{H}}$ is all histories in $\mathcal{H}$ that are boundary consistent. We rely on this notion of boundary consistency in generalizing the Shapley axioms to any explanation boundary, $\mathcal{B}$:

\begin{itemize}
    \item Efficiency: \small{$\sum_{i \in cut(\mathcal{B})} \phi_{\nu_{\mathcal{B}}}(i) =  f(\mathbf{x}) - f(\mathbf{x'})$}. 
    
    In the general case where $\nu_{\mathcal{B}}$ can depend on the ordering of $h$, the sum is $\sum_{h \in \tilde{\mathcal{H}}_{\mathcal{B}}} \frac{\nu_{\mathcal{B}}(h)}{ |\tilde{\mathcal{H}}_{\mathcal{B}}|} - \nu_{\mathcal{B}}([])$. But when the game is defined by a model function $f$, $\sum_{h \in \tilde{\mathcal{H}}_{\mathcal{B}}} \nu_{\mathcal{B}}(h) / |\tilde{\mathcal{H}}_{\mathcal{B}}|=f(\boldsymbol{x})$ and
    $\nu_{\mathcal{B}}([])=f(\boldsymbol{x'})$. An illustration with $3$ boundaries is shown in \textbf{Figure \ref{fig:efficiency_and_boundary_consistency}}.
    
    \item Linearity: $\phi_{\alpha u + \beta v} = \alpha \phi_{u} + \beta \phi_{v}$ for any payoff functions $u$ and $v$ and scalars $\alpha$ and $\beta$.   
    
    Linearity enables us to compute a linear ensemble of models by independently explaining each model and then linearly weighting the attributions. Similarly, we can explain $f(\mathbf{x}) - \mathbb{E}(f(X'))$ by independently computing attributions for each background sample $\mathbf{x^{(i)'}}$ and then taking the average of the attributions, without recomputing from scratch whenever the background sample's distribution changes. An illustration with $2$ background samples is shown in \textbf{Figure \ref{fig:linearity}}.  
    % Quoting from \cite{frye2019asymmetric}, linearity ensures that attribution ``for a linear ensemble model are the linear combination of" attributions ``for the members of the ensemble".
    
    \item Dummy player: $\phi_{\nu_{\mathcal{B}}}(i) = 0$ if
    $\nu_{\mathcal{B}}(S+[i] )=\nu_{\mathcal{B}}(S)$ 
    for all  $S, S+[i] \in \tilde{\mathcal{C}}_{\mathcal{B}}$ for $i \in cut(\mathcal{B})$.
    
    Dummy player states that if an edge does not change the model's output when added to in all possible coalitions, it should be given $0$ attribution. In \textbf{Figure \ref{fig:dummy}}, $e_2$ is a dummy edge because starting from any coalition, adding $e_2$ wouldn't change the output.
    
\end{itemize}

These last three axioms are extensions of Shapley's axioms. Note that Shapley value also requires the symmetry axiom because the game is defined on a set of players. For Shapley Flow values this symmetry assumption is encoded through our choice of an ordered history formulation. %encoded Since order is independent in a set, symmetry is required to uniquely attribute credit to players. However, we do not require the symmetry axiom because our game is defined on histories, which results in a natural  asymmetry between players. A detailed discussion is included in 
(Appendix \textbf{Section \ref{sec:proof}}). 

\subsection{Shapley Flow is the unique solution}

Shapley Flow uniquely satisfies all axioms from the previous section. Here, we describe the algorithm, show its formulae, and state its properties. Please refer to \textbf{Appendix \ref{sec:alg}} and \textbf{\ref{sec:proof}} for the pseudo code\footnote{code can be found in \url{https://github.com/nathanwang000/Shapley-Flow}} and proof.

\textbf{Description}: Define a configuration of a graph as an arbitrary ordering of outgoing edges of a node when it is traversed by depth first search. For each configuration, we run depth first search starting from the source node, processing edges in the order of the configuration. When processing an edge, we update the value of the edge’s target node by making the edge’s source node value visible to its function. If the edge’s target node is the sink node, the difference in the sink node's output is credited to every edge along the search path from source to sink. The final result averages over attributions for all configurations. 

\textbf{Formulae}: Denote the attribution of Shapley Flow to a path as $\tilde{\phi}_\nu$, and the set of all possible orderings of source nodes to a sink path generated by depth first search (DFS) as $\Pi_{\text{dfs}}$. For each ordering $\pi \in \Pi_{\text{dfs}}$, the inequality of $\pi(j) < \pi(i)$ denotes that path $j$ precedes path $i$ under $\pi$. Since $\nu$'s input is a list of edges, we define $\tilde{\nu}$ to work on a list of paths. The evaluation of $\tilde{\nu}$ on a list of paths is the value of $v$ evaluated on the corresponding edge traversal ordering.  Then

\begin{equation} \label{eq:dfs}
    \tilde{\phi}_{\nu}(i) = \sum_{\pi \in \Pi_{\text{dfs}}} \frac{\tilde{\nu}([j: \pi(j) \leq \pi(i)]) - \tilde{\nu}([j: \pi(j) < \pi(i)])}{|\Pi_{\text{dfs}}|}
\end{equation}

To obtain an edge $e$'s attribution $\phi_v(e)$, we sum the path attributions for all paths that contains $e$.
\begin{equation}
    \phi_{\nu}(e) = \sum_{p \in \text{paths in } \mathcal{G}}
    \mathds{1}_{p \text{ contains}}(e) \tilde{\phi}_{\nu}(p)
\end{equation}

\textbf{Additional properties}: Shapley Flow has the following beneficial properties beyond the axioms.

    \underline{Generalization of SHAP}: if the graph is flat, the edge attribution is equal to feature attribution from SHAP because each input node is paired with a single edge leading to the model.
    
    \underline{Generalization of ASV}: the attribution to the source nodes is the same as in ASV if all the dependencies among features are modeled by the causal graph.
    
    \underline{Generalization of Owen value}: if the graph is a tree, the edge attribution for incoming edges to the leaf nodes is the Owen value \citep{lopez2009relationship} with a coalition structure defined by the tree. 
    
    \underline{Implementation invariance}: implementation invariance means that no matter how the function is implemented, so long as the input and output remain unchanged, so does the attribution \citep{sundararajan2017axiomatic}, which directly follows boundary consistency (\textit{i.e.}, knowing $f$'s computational graph or not wouldn't change the upstream attribution).
    
    \underline{Conservation of flow}: efficiency and boundary consistency imply that the sum of attributions on a node's incoming edges equals the sum of its outgoing edges.
    
    \underline{Model agnostic}: Shapley Flow can explain arbitrary (non-differentiable) machine learning pipelines.

\section{Practical Application}

Shapley Flow highlights both the direct and indirect impact of features. In this section, we consider several applications of Shapley Flow. First, in the context of a linear model, we verify that the attributions match our intuition. Second, we show how current feature attribution approaches lead to an incomplete understanding of a system compared to Shapley Flow.

\subsection{Experimental Setup}

We illustrate the application of Shapley Flow to a synthetic and a real dataset.
%, as it have been widely used in the feature attribution literature \citep{lundberg2020local}
In addition, we include results for a third dataset in the Appendix. Note that our algorithm assumes a causal graph is provided as input. In recent years there has been significant progress in causal graph estimation \citep{glymour2019review,peters2017elements}. However, since our focus is not on causal inference, we make simplifying assumptions in estimating the causal graphs (see \textbf{Section \ref{sec:causal_detail}} of the Appendix for details). 

\textbf{Datasets.} 
\textit{Synthetic}: As a sanity check, we first experiment with  synthetic data. We create a random graph dataset with $10$ nodes. %, labeling each node from $1$ to $10$. 
A node $i$ is randomly connected to node $j$ (with $j$ pointing to $i$) with $0.5$ probability if $i > j$, otherwise $0$. The function at each node is linear with weights generated from a standard normal distribution. Sources follow a $N(0,1)$ distribution. This results in a graph with a single sink node associated with function $f$ (\textit{i.e.}, the `model' of interest). The remainder of the graph corresponds to the causal structure among the input variables.% This ensures the resulting graph is a DAG. The weights on the connection is generated from a standard normal distribution. Assuming the source nodes of this random graph follows standard normal distribution, we sample $1000$ examples to explain.

\textit{National Health and Nutrition Examination Survey}: This dataset consists of $9,932$ individuals with $18$ demographic and laboratory measurements \citep{cox1998plan}. We used the same preprocessing as described by \cite{lundberg2020local}.  Given these inputs, the model, $f$, aims to predict survival. 

\textbf{Model training.} We train $f$ using an $80/20$ random train/test split. For experiments with linear models, $f$ is trained with linear regression. For experiments with non-linear models, $f$ is fitted by $100$ XGBoost trees with a max depth of $3$ for up to $1000$ epochs, using the Cox loss.
 %For the nutrition dataset, $f$ is trained with cox loss.

\textbf{Causal Graph.} For the nutrition dataset, we constructed a causal graph (\textbf{Figure \ref{fig:nutrition}}) based on our limited understanding of the causal relationship among input variables. This graph represents an oversimplification of the true underlying causal relationships and is for illustration purposes only. 
%Our causal graph comes from...  
We assigned attributes predetermined at birth (age, race, and sex) as source nodes because they temporally precede all other features. Poverty index depends on age, race, and sex (among other variables captured by the poverty index noise variable) and impacts one's health. Other features pertaining to health depend on age, race, sex, and poverty index. Note that the relationship among some features is deterministic. For example, pulse pressure is the difference between systolic and diastolic blood pressure. We include causal edges to account for such facts. We also account for when features have natural groupings. For example, transferrin saturation (TS), total iron binding capacity (TIBC), and serum iron are all related to blood iron. Serum albumin and serum protein are both blood protein measures. Systolic and diastolic blood pressure can be grouped into blood pressure. Sedimentation rate and white blood cell counts both measure inflammation. We add these higher level grouping concepts as new latent variables in the graph. To account for noise in modeling the outcome (\textit{i.e.}, the effect of exogenous variables that are not used as input to the model), we add an independent noise node to each node (detailed in \textbf{Section \ref{sec:causal_detail}} in the Appendix). \textbf{The resulting causal structure is an oversimplification of the true causal structure; the relationship between source nodes (e.g., race) and biomarkers is far more complex \citep{robinson2020teaching}. Nonetheless, it can help in understanding the in/direct effects of input variables on the outcome}. %As a result, not all nodes have a direct edge to the model. 

\subsection{Baselines}

We compare Shapley Flow with other game theoretic feature attribution methods: independent SHAP \citep{lundberg2017unified}, on-manifold SHAP \citep{aas2019explaining}, and ASV \citep{frye2019asymmetric}, covering both independent and on-manifold feature attribution.

%\subsection{Implementation Details}

Since Shapley value based methods are expensive to compute exactly, we use a Monte Carlo approximation of \textbf{Equation \ref{eq:dfs}}. In particular, we sample orderings from $\Pi_{\text{dfs}}$ and average across those orderings.% in accordance with \cite{frye2019asymmetric}. 
 We randomly selected a background sample from each dataset and share it across methods so that each uses the same background. A single background sample allows us to ignore differences in methods due to variations in background sampling and is easier to explain the behavior of baselines \citep{merrick2020}. To show that our result is not dependent on the particular choice of background sample, we include an example averaged over $100$ background samples in \textbf{Section \ref{sec:multiple_background}} in the Appendix (the qualitative results shown with a single background still holds). We sample $10,000$ orderings from each approach to generate the results. Since there's no publicly available implementation for ASV, we show the attribution for source nodes (the noise node associated with each feature) obtained from Shapley Flow (summing attributions of outgoing edges), as they are equivalent given the same causal graph. Since noise node's credit is used, intermediate nodes can report non zero credit in ASV.

 For convenience of visual inspection, we show top $10$ links used by Shapley Flow (credit measured in absolute value) on the nutrition dataset.

\subsection{Sanity checks with linear models}\label{sec:sanity_linear}

To build intuition, we first examine linear models (\textit{i.e.}, $f(\mathbf{x}) = \mathbf{w}^\top \mathbf{x} + b$ where $\mathbf{w} \in \mathbb{R}^d $ and $b \in \mathbb{R}$; the causal dependence inside the graph is also linear). When using a linear model the ground truth direct impact of changing feature $X_i$ is $w_i (x_i - x_i')$ (that is the change in output due to $X_i$ directly), and the ground truth indirect impact is defined as the change in output when an intervention changes  $x_i'$ to $x_i$. Note that when the model is linear, only $1$ Monte Carlo sample is sufficient to recover the exact attribution because feature ordering doesn't matter (the output function is linear in any boundary edges, thus only the background and foreground value of a feature matters). This allows us to bypass sampling errors and focus on analyzing the algorithms.

 Results for explaining the datasets are included in \textbf{Table \ref{tab:direct_and_indirect_effect}}. We report the mean absolute error (and its variance) associated with the estimated attribution (compared against the ground truth attribution), averaged across $1,000$ randomly selected test examples and all graph nodes for both datasets. Note that only Shapley flow results in no error for both direct and indirect effects.

%We generate random linear models by sampling $\mathbf{w} \sim N(0,I)$ for every node's function. We generate $1000$ examples by independently sampling the source nodes from a standard gaussian distribution. With a randomly generated graph, we compute the difference in feature importance with ground truth across $1000$ examples. Since Shapley Flow attributes to edges, we use the attribution of direct edges pointing to $f$ as its source node's direct feature attribution, and use the sum of a node's outgoing edge attribution as its indirect feature attribution. We repeat the same experiment for the income and nutrition datasets (fitting linear models). As shown in \textbf{Table \ref{tab:direct_effect}}, Shapley Flow and independent SHAP make no mistake for the direct effect.
%However, only Shapley Flow shows no mistake for indirect effect (ASV fails on non-source nodes) in \textbf{Table \ref{tab:indirect_effect}}.

\begin{table}
\scalebox{0.65}{
\begin{tabular}{lrrrrr}
\toprule
{Methods} &  Nutrition (\textbf{D})  &  Synthetic (\textbf{D}) & Nutrition (\textbf{I})  &  Synthetic (\textbf{I}) \\
\midrule
Independent & {\bf 0.0} ($\pm$ 0.0) & {\bf0.0} ($\pm$ 0.0) & 0.8 ($\pm$ 2.7) & 1.1 ($\pm$ 1.4) \\
On-manifold & 1.3 ($\pm$ 2.5) &   0.8 ($\pm$ 0.7) & 0.9 ($\pm$ 1.6) &   1.5 ($\pm$ 1.5)\\
ASV & 1.5 ($\pm$ 3.3) & 1.2 ($\pm$ 1.4) & 0.6 ($\pm$ 1.9) & 1.1 ($\pm$ 1.5) \\
Shapley Flow & {\bf 0.0} ($\pm$ 0.0) & {\bf0.0} ($\pm$ 0.0) & {\bf 0.0} ($\pm$ 0.0) & {\bf0.0} ($\pm$ 0.0)\\
\bottomrule
\end{tabular}
}
\setlength{\belowcaptionskip}{-10pt}
\caption{Mean absolute error (std) for all methods on direct (\textbf{D}) and indirect (\textbf{I}) effect for linear models. Shapley Flow makes no mistake across the board. %For comparison, the standard deviation in output is 1.6 for the income dataset, 2.7 for the nutrition dataset, and 3.1 for the synthetic dataset
}
\label{tab:direct_and_indirect_effect}
\end{table}

% \begin{figure}
%     \centering

%     \subfloat[Synthetic direct effect]{\label{fig:synth_direct_error}
%     \includegraphics[width=0.45 \linewidth]{figs/synthetic_direct_error.png}}
%     \subfloat[Synthetic indirect effect]{\label{fig:synth_indirect_error}
%     \includegraphics[width=0.45\linewidth]{figs/synthetic_indirect_error.png}}
    
%     \caption{Linear synthetic dataset shows that only Shapley Flow shows the direct and indirect effect of features.}
% \label{fig:linear_synthetic_viz}
% \end{figure}

% \begin{figure}
%     \centering

%     \subfloat[Income direct effect]{\label{fig:income_direct_error}
%     \includegraphics[width=0.45 \linewidth]{figs/income_direct_error.png}}
%     \subfloat[Income indirect effect]{\label{fig:income_indirect_error}
%     \includegraphics[width=0.45\linewidth]{figs/income_indirect_error.png}}

%     \subfloat[Nutrition direct effect]{\label{fig:nutrition_direct_error}
%     \includegraphics[width=0.45 \linewidth]{figs/nutrition_direct_error.png}}
%     \subfloat[Nutrition indirect effect]{\label{fig:nutrition_indirect_error}
%     \includegraphics[width=0.45\linewidth]{figs/nutrition_indirect_error.png}}
    
%     \caption{Linear models on Income and Nutrition datasets. Only Shapley Flow shows the direct and indirect effect of features.}
% \label{fig:linear_real_viz}
% \end{figure}

\subsection{Examples with non-linear models}

We demonstrate the benefits of Shapley Flow with non-linear models containing both discrete and continuous variables. As a reminder, the baseline methods are not competing with Shapley Flow as the latter can recover all the baselines given the corresponding causal structure (\textbf{Figure \ref{fig:pearl_example}}). Instead, we highlight why a holistic understanding of the system is better.

\begin{figure}
    \centering
    \scalebox{0.8}{
\begin{tabular}{lrrr}
\toprule
{Top features} &   Age &  Serum Magnesium &  Serum Protein \\
\midrule
Background sample  &  35 &             1.37 &   7.6 \\
Foreground sample &  40 &             1.19 &   6.5 \\
\bottomrule
\end{tabular}
}

\centering

\scalebox{0.85}{
\begin{tabular}{lrrr}
\toprule
{Attributions} & Independent & On-manifold & ASV\\
\midrule
Age & \cellcolor{red!100} 0.1 & \cellcolor{blue!100} -0.26 & \cellcolor{red!100} 0.16\\
Serum Magnesium & \cellcolor{red!21} 0.02 & \cellcolor{red!78} 0.2 & \cellcolor{red!15} 0.02\\
Serum Protein & \cellcolor{blue!81} -0.09 & \cellcolor{red!26} 0.07 & \cellcolor{red!0} 0.0\\
Blood pressure & \cellcolor{red!0} 0.0 & \cellcolor{red!0} 0.0 & \cellcolor{blue!92} -0.14\\
Systolic BP & \cellcolor{blue!50} -0.05 & \cellcolor{blue!19} -0.05 & \cellcolor{red!0} 0.0\\
Diastolic BP & \cellcolor{blue!38} -0.04 & \cellcolor{blue!25} -0.07 & \cellcolor{red!0} 0.0\\
Serum Cholesterol & \cellcolor{red!0} 0.0 & \cellcolor{blue!57} -0.15 & \cellcolor{red!0} 0.0\\
Serum Albumin & \cellcolor{red!2} 0.0 & \cellcolor{blue!54} -0.14 & \cellcolor{red!0} 0.0\\
Blood protein & \cellcolor{red!0} 0.0 & \cellcolor{red!0} 0.0 & \cellcolor{blue!53} -0.08\\
White blood cells & \cellcolor{red!0} 0.0 & \cellcolor{red!44} 0.11 & \cellcolor{red!0} 0.0\\
Race & \cellcolor{red!0} 0.0 & \cellcolor{red!33} 0.09 & \cellcolor{red!0} 0.0\\
BMI & \cellcolor{red!0} -0.0 & \cellcolor{red!29} 0.08 & \cellcolor{red!0} -0.0\\
TIBC & \cellcolor{red!1} 0.0 & \cellcolor{red!24} 0.06 & \cellcolor{red!0} 0.0\\
Sex & \cellcolor{red!0} 0.0 & \cellcolor{blue!20} -0.05 & \cellcolor{red!0} 0.0\\
TS & \cellcolor{red!0} 0.0 & \cellcolor{red!18} 0.05 & \cellcolor{red!0} 0.0\\
Pulse pressure & \cellcolor{red!0} 0.0 & \cellcolor{blue!17} -0.05 & \cellcolor{red!0} 0.0\\
Poverty index & \cellcolor{red!1} 0.0 & \cellcolor{red!15} 0.04 & \cellcolor{red!0} 0.0\\
Red blood cells & \cellcolor{red!0} 0.0 & \cellcolor{red!10} 0.03 & \cellcolor{red!0} 0.0\\
Serum Iron & \cellcolor{red!0} 0.0 & \cellcolor{blue!7} -0.02 & \cellcolor{red!0} 0.0\\
Sedimentation rate & \cellcolor{red!0} 0.0 & \cellcolor{red!1} 0.0 & \cellcolor{red!0} 0.0\\
Iron & \cellcolor{red!0} 0.0 & \cellcolor{red!0} 0.0 & \cellcolor{red!0} -0.0\\
Inflamation & \cellcolor{red!0} 0.0 & \cellcolor{red!0} 0.0 & \cellcolor{red!0} 0.0\\
\bottomrule
\end{tabular}
}

\subfloat[Shapley Flow]{\label{fig:nutrition_flow_38}
    \includegraphics[width=0.9\linewidth]{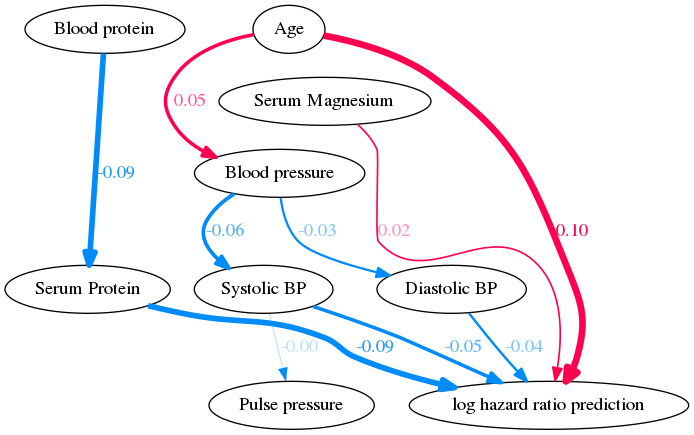}}
    
    \setlength{\belowcaptionskip}{-20pt}
    \caption{Comparison among baselines on a sample (top table) from the nutrition dataset, showing top 10 features/edges.}
\label{fig:nutrition_viz_38}
\end{figure}

\textbf{Independent SHAP ignores the indirect impact of features}. Take an example from the nutrition dataset (\textbf{Figure  \ref{fig:nutrition_viz_38}}). Independent SHAP gives lower attribution to age compared to ASV. This happens because age, in addition to its direct impact, indirectly affects the output through blood pressure, as shown by Shapley Flow (\textbf{Figure \ref{fig:nutrition_flow_38}}). %In particular, race partially accounts for the impact of serum magnesium because changing race from Black to White on average increases serum magnesium by $0.07$ meg/L in the dataset (thus partially explaining the increase in serum magnesium changing from the background sample to the foreground). 
Independent SHAP fails to account for the indirect impact of age, leaving the user with a potentially misleading impression that age is less important than it actually is.

\begin{figure}
    \centering
    
    \subfloat[Age vs. output]{\label{fig:nutrition_age_38}
    \includegraphics[width=0.48 \linewidth]{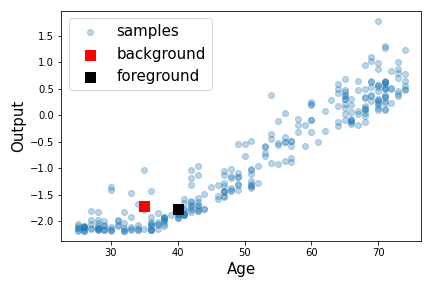}}
    \subfloat[Age vs. magnesium]{\label{fig:nutrition_age_serum_magnesium_38}
    \includegraphics[width=0.48\linewidth]{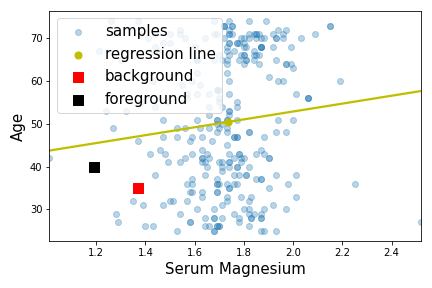}}
    \setlength{\belowcaptionskip}{-10pt}

    \caption{Age appears to be protective in on-manifold SHAP because it steals credit from other variables.}
\label{fig:nutrition_explain_on_manifold_38}
\end{figure}

\textbf{On-manifold SHAP provides a misleading interpretation}. With the same example (\textbf{Figure  \ref{fig:nutrition_viz_38}}), we observe that on-manifold SHAP strongly disagrees with independent SHAP, ASV, and Shapley Flow on the importance of age. Not only does it assign more credit to age, it also flips the sign, suggesting that age is protective. However, \textbf{Figure \ref{fig:nutrition_age_38}} shows that age and earlier mortality are positively correlated; then how could age be protective?  \textbf{Figure \ref{fig:nutrition_age_serum_magnesium_38}} provides an explanation. Since SHAP considers all partial histories regardless of the causal structure, when we focus on serum magnesium and age, there are two cases: serum magnesium updates before or after age. We focus on the first case because it is where on-manifold SHAP differs from other baselines (all baselines already consider the second case as it satisfies the causal ordering). When serum magnesium updates before age, the expected age given serum magnesium is higher than the foreground age (yellow line above the black marker). Therefore when age updates to its foreground value, we observe a decrease in age, leading to a decrease in the output (so age appears to be protective). %Serum magnesium is just one variable from which age steals credit. Similar logic applies to  TIBC, red blood cells, serum protein, serum cholesterol, and diastolic BP. 
From both an in/direct impact perspective, on-manifold perturbation can be misleading since it is based not on causal but on observational relationships.

\textbf{ASV ignores the direct impact of features}. As shown in \textbf{Figure \ref{fig:nutrition_viz_38}}, serum protein appears to be more important in independent SHAP compared to ASV. From Shapley Flow (\textbf{Figure \ref{fig:nutrition_flow_38}}), we know serum protein is not given attribution in ASV because its upstream node, blood protein, gets all the credit. However, looking at ASV alone, one fails to understand that intervening on serum protein could have a larger impact on the output. 

\textbf{Shapley Flow shows both direct and indirect impacts of features}. Focusing on the attribution given by Shapley Flow (\textbf{Figure \ref{fig:nutrition_flow_38}}). We not only observe similar direct impacts in variables compared to independent SHAP, but also can trace those impacts to their source nodes, similar to ASV. Furthermore, Shapley Flow provides more detail compared to other approaches. For example, using Shapley Flow we gain a better understanding of the ways in which age impacts survival. The same goes for all other features. This is useful because causal links can change (or break) over time. Our method provides a way to reason through the impact of such a change.

More case studies with an additional dataset are included in the Appendix.
%We provide more case studies and an additional dataset highlighting the utility of Shapley Flow in the Appendix. %In addition to understanding a single example, we also show how Shapley Flow can be used to understand an entire dataset. Please refer to the Appendix for more details.

\section{Discussion and Conclusion}

We extend the classic Shapley value axioms to causal graphs, resulting in a unique edge attribution method: Shapley Flow. It unifies three previous Shapley value based feature attribution methods, and enables the joint understanding of both the direct and indirect impact of features. This more comprehensive understanding is useful when interpreting any machine learning model, both `black box' methods, and `interpretable' methods (such as linear models). 

The key message of the paper is that model interpretation methods should include the whole machine learning pipeline in order to understand when a model can be applied. While our approach relies on access to a complete causal graph, Shapley Flow is still valuable because a) there are well-established causal relationships in domains such as healthcare, and ignoring such relationships can produce confusing explanations; b) recent advancements in causal estimation are complementary to our work and make defining these graphs easier; c) finally and most importantly, existing methods already implicitly make causal assumptions, Shapley Flow just makes these assumptions explicit (\textbf{Figure \ref{fig:pearl_example}}). However, this does open up new research opportunities. Can Shapley Flow work with partially defined causal graphs? How to explore Shapley Flow attribution when the causal graph is complex? Can Shapley Flow be useful for feature selection? We leave those questions for future work.

% \textit{The causal graph gives Shapley Flow an advantage in the experiments.} Yes, but a) both ASV and Shapley Flow have access to the causal graph, and b) we are specifically considering scenarios where an estimated causal graph is available, so it is desirable to use it rather than ignore it.

% \textbf{Even with interpretable models, the problems persist:} The highlighted problems are not restricted to black-box models. Even if the model and the causal structure are easy to interpret (\textit{e.g.}, all are linear models), feature attribution with a fixed boundary is still flawed. %Consider we have linear causal equations and a linear model. For any fixed boundary, message arrival order doesn't matter. This greatly simplifies the computation because we just need to sample one DFS ordering. {\color{red} discussion of what should I show because this point seems obvious to me}

\bibliography{main}

\newpage
\onecolumn

%\section{Appendix}
\section{Explanation boundary for on-manifold methods without a causal graph}\label{sec:on_manifold_boundary}

\begin{figure}%[H]
\centering
\subfloat[On-manifold attribution]{\label{fig:illustration_on_manifold}
\includegraphics[width=0.35\linewidth]{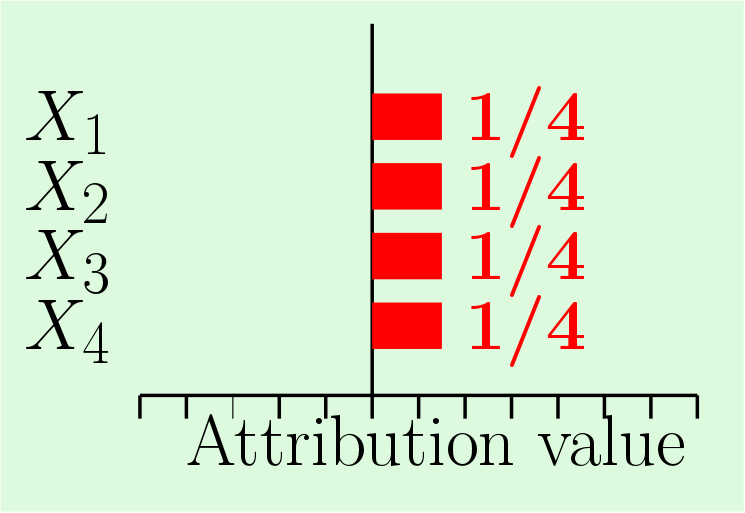}
}
\subfloat[On-manifold boundary]{\label{fig:boundary_on_manifold}
\includegraphics[width=0.35\linewidth]{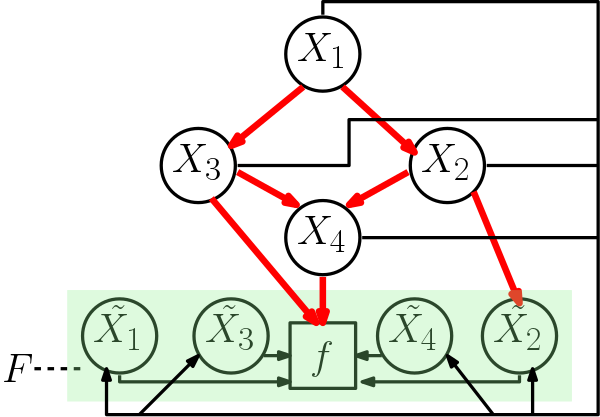}
}
\setlength{\belowcaptionskip}{-14pt}
\caption{On manifold perturbation methods can be computed using Shapley Flow with a specific explanation boundary.}
\label{fig:pearl_example_on_manifold}
\end{figure}

On-manifold perturbation using conditional expectations can be unified with Shapley Flow using explanation boundaries (\textbf{Figure \ref{fig:illustration_on_manifold}}). Here we introduce $\tilde{X_i}$ as an auxiliary variable that represent the imputed version of $X_i$. Perturbing any feature  $X_i$ affects all input to the model ($\tilde{X_1}$, $\tilde{X_2}$, $\tilde{X_3}$, $\tilde{X_4}$) so that they respect the correlation in the data after the perturbation. When $X_i$ has not been perturbed, $\tilde{X}_j$ treats it as missing for $i, j \in [1,2,3,4]$ and would sample $\tilde{X}_j$ from the conditional distribution of $X_j$ given non-missing predecessors. The red edges contain causal links from \textbf{Figure \ref{fig:pearl_causal_graph}}, whereas the black edges are the causal structure used by the on-manifold perturbation method. The credit is equally split among the features because they are all correlated. Again, although giving $X_1$ and $X_2$ credit is not true to $f$, it is true to the model defined by $F$. 

\section{The Shapley Flow algorithm} \label{sec:alg}

A pseudo code implementation highlighting the main ideas for Shapley Flow is included in \textbf{Algorithm \ref{alg:shapley_flow}}. %An exact implementation is in \textbf{Algorithm \ref{alg:shapley_flow_detail}}. 
For approximations, instead of trying all edge orderings in line $15$ of \textbf{Algorithm \ref{alg:shapley_flow}}, one can try random orderings and average over the number of orderings tried. 

\begin{algorithm}
  \caption{Shapley Flow pseudo code %(detailed implementation is in {\bf Algorithm \ref{alg:shapley_flow_detail}})
    \label{alg:shapley_flow}
    }
    \textbf{Input:} A computational graph $\mathcal{G}$ (each node $i$ has a function $f_i$), foreground sample $\mathbf{x}$, background sample $\mathbf{x'}$ \\
    \textbf{Output:} Edge attribution $\phi: E \rightarrow \mathbb{R}$ \\
    \textbf{Initialization:} 
    \\$\mathcal{G}$: add an new source node pointing to original source nodes. 
  \begin{algorithmic}[1]
    %\Statexgithub
    
    \Function{ShapleyFlow}{$\mathcal{G}$, $\mathbf{x}'$, $\mathbf{x}$}
    
        \State \Call{Initialize}{$\mathcal{G}$, $\mathbf{x}'$, $\mathbf{x}$} \Comment{Set up game $\nu$ for any boundary in $\mathcal{G}$}
        \Let{$s$}{\Call{source}{$\mathcal{G}$}} \Comment{Obtain the source node}
        \State \Return{\Call{DFS}{$s$, $\{\}$, $[]$}}
    \EndFunction
    
    \Statex
    \Function{DFS}{$s$, $D$, $S$}
        \LineComment{$s$ is a node, $D$ is the data side of the current boundary, $S$ is coalition}
        \LineComment{Using Python list slice notation}
        \State Initialize $\phi$ to output $0$ for all edges
        
        \If{\Call{IsSinkNode}{s}}
        \LineComment{Here we overload $D$ to refer to its boundary} %For convenience of notation, we overload the meaning of $F$ to also refer to the computation represented by nodes in $F$. Since $F$ is associated with $\mathcal{B}$ in a one to one manner, we use them interchangeably to refer to a boundary in figure illustrations.}
            \Let{$\phi(S[-1])$}{$\nu_{D}(S) - \nu_{D}(S[:-1])$} \Comment{Difference in output is attributed to the edge}
            \State \Return{$\phi$}
        \EndIf
            
        % \Statex
        % \Let{state}{$\{\}$} \Comment{Initialize state to an empty dictionary}
        % \State \Call{SaveState}{$s$, state}
        
        \Statex
        \For{$p \gets \text{AllOrderings}(\text{Children}(s))$} \Comment{Try all orderings/permutations of the node's children}
        
            %\State \Call{LoadState}{$s$, state} \Comment{Restore settings of nodes' values}
            
            \For{$c \gets p$} \Comment{Follow the permutation to get the node one by one}
                \Let{edgeCredit}{\Call{DFS}{$c$, $D \cup \{s\}$, $S + [(s, c)]$}} \Comment{Recurse downward}
                
                \Statex
                \Let{$\phi$}{$\phi + \frac{\text{edgeCredit}}{\text{NumChildren}(s)!}$} \Comment{Average attribution over number of runs}
                \Let{$\phi(S[-1])$}{$\phi(S[-1]) + \frac{\text{edgeCredit}(s, c)}{\text{NumChildren}(s)!}$} \Comment{Propagate upward}
            \EndFor
        
        \EndFor
        \State \Return{$\phi$}
    
    \EndFunction
    
  \end{algorithmic}
\end{algorithm}

\section{Shapley Flow's uniqueness proof} \label{sec:proof}

Without loss of generality, we can assume $\mathcal{G}$ has a single source node $s$. We can do this because every node in a causal graph is associated with an independent noise node \cite[Chapter~6]{peters2017elements}. For deterministic relationships, the function for a node doesn't depend on its noise. Treating those noise nodes as a single node, $s$, wouldn't have changed any boundaries that already exist in the original graph. Therefore we can assume there is a single source node $s$.

\subsection{At most one solution satisfies the axioms}

Assuming that a solution exists, we show that it must be unique. %The idea is to ``grow" the explanation boundary one node at a time, maintaining the property that credit of previous edges are unique. % Then, following the standard proof for uniqueness of Shapley value, we show that every game at each boundary level can be uniquely factored into a linear combination of basis games, where each basis game has a unique attribution, thus completing the proof. 

\begin{proof}

We adapt the argument from the Shapley value uniqueness proof
\footnote{\url{https://ocw.mit.edu/courses/economics/14-126-game-theory-spring-2016/lecture-notes/MIT14_126S16_cooperative.pdf}}, by defining basis payoff functions as carrier games. Choose any boundary $\mathcal{B}$, we show here that any game defined on the boundary has a unique attribution. We also drop the subscript $\mathcal{B}$ in the proof as there is no ambiguity. Note that since every edge will appear in some boundary, if all boundary edges are uniquely attributed to, all edges have unique attributions. A carrier game associated with coalition (ordered list) $O$ is a game with payoff function $v^O$ such that $v^O(S) = 1(0)$ if coalition $S$ starts with $O$ (otherwise 0). By dummy player, we know that only the last edge $e$ in $O$ gets credit and all other edges in the cut set are dummy because a coalition is constructed in order (only adding $e$ changes the payoff from $0$ to $1$). Note that in contrast with the traditional symmetry axiom \citep{shapley1953value} defined on a set of players, the symmetry axiom is not explicit in our case (it is made implicitly) because not all edges in the carrier game are symmetric with each other (observe that $e$ is different from all other edges, which are dummy), thus we do not need an explicit symmetry axiom to argue for unique attribution in the carrier game. Furthermore, $e$ must be an edge in the boundary to form a valid game because boundary edges are the only edges that are connected to the model defined by the boundary. Therefore we give $0$ credit to edges in the cut set other than $e$ (because they are \textit{dummy players}). By the \textit{efficiency axiom}, we give $\sum_{h \in \tilde{\mathcal{H}}} \frac{\nu_{\mathcal{B}}(h)}{ |\tilde{\mathcal{H}}|} - \nu_{\mathcal{B}}([])$ credit to $e$ where $\tilde{\mathcal{H}}$ is the set of all possible boundary consistent histories as defined in \textbf{Section \ref{sec:axioms}}. This uniquely attributed the boundary edges for this game.

We show that the set of carrier games associated with every coalition that ends in a boundary edge (denoted as $\hat{\mathcal{C}}$) form basis functions for all payoff functions associated with the system. Recall from \textbf{Section \ref{sec:game}} that $\tilde{\mathcal{C}}$ is the set of \textit{boundary consistent coalitions}. We show here that payoff value on coalitions from $\tilde{\mathcal{C}}$ is redundant given $\hat{\mathcal{C}}$. Note that $\tilde{\mathcal{C}}$ \textbackslash $\hat{\mathcal{C}}$ represents all the coalitions that do not end in a boundary edge. For $c \in \tilde{\mathcal{C}}$ \textbackslash $\hat{\mathcal{C}}$, $v^O(c) = v^O(c[:-1])$ (using Python's slice notation on list) because only boundary edges are connected to the model defined by the boundary. %Therefore the payoff value for every coalition in $\tilde{\mathcal{C}}$ is determined by what is in $v^O(\hat{\mathcal{C}})$. %First note that $\hat{\mathcal{C}}$ has the same cardinality as the space of all payoff functions associated with a graph because dimensions for $\tilde{\mathcal{C}}$ \textbackslash $\hat{\mathcal{C}}$ are completely determined by $\hat{\mathcal{C}}$ (there are redundant dimensions in the input space of all payoff functions because non boundary edges have no impact). 
Therefore it suffices to show that $v^O$ is linearly independent for $O \in \hat{\mathcal{C}}$. For a contradiction, assume for all $c \in \hat{\mathcal{C}}$, $\sum_{O \subseteq \hat{\mathcal{C}}} \alpha^O v^O(c) = 0$, with some non zero $\alpha^O \in \mathbb{R}$ (definition of linear dependence). Let $S$ be a coalition with minimal length such that $\alpha^S \neq 0$. We have $\sum_{O \subseteq \hat{\mathcal{C}}} \alpha^O v^O(S) = \alpha^S$, a contradiction.

Therefore for any $\nu$ we have unique $\alpha$'s such that $\nu = \sum_{O \subseteq \hat{\mathcal{C}}} \alpha^O v^O$. Using the \textit{linearity axiom}, we have 

$$\phi_{\nu} = \phi_{\sum_{O \subseteq \hat{\mathcal{C}}}  \alpha^O v^O} = \sum_{O \subseteq \hat{\mathcal{C}}} \alpha^O \phi_{v^O}$$

The uniqueness of $\alpha$ and $\phi_{v^O}$ makes the attribution unique if a solution exists. Axioms used in the proof are italicized.

\end{proof}

\subsection{Shapley Flow satisfies the axioms}

\begin{proof}

We first demonstrate how to generate all boundaries. Then we show that Shapley Flow gives boundary consistent attributions. Following that, we look at the set of histories that can be generated by DFS in boundary $\mathcal{B}$, denoted as $\Pi^{\text{dfs}}_{\mathcal{B}}$. We show that $\Pi^{\text{dfs}}_{\mathcal{B}}=\tilde{\mathcal{H}}_{\mathcal{B}}$. Using this fact, we check the axioms one by one.

\begin{itemize}
\item Every boundary can be ``grown" one node at a time from $D=\{s\}$ where $s$ is the source node: Since the computational graph $\mathcal{G}$ is a directed acyclic graph (DAG), we can obtain a topological ordering of the nodes in $\mathcal{G}$. Starting by including the first node in the ordering (the source node $s$), which defines a boundary as $(D=\{s\}, F=\text{Nodes}(\mathcal{G}) \backslash D)$, we grow the boundary by adding nodes to $D$ (removing nodes from $F$) one by one following the topological ordering. This ordering ensures the corresponding explanation boundary is valid because the cut set only flows from $D$ to $F$ (if that's not true, then one of the dependency nodes is not in $D$, which violates topological ordering). 

Now we show every boundary can be ``grown" in this fashion. In other words, starting from an arbitrary boundary $\mathcal{B}_1 = (D_1, F_1)$, we can ``shrink" one node at a time to $D=\{s\}$ by reversing the growing procedure. First note that, $D_1$ must have a node with outgoing edges only pointing to nodes in $F_1$ (if that's not the case, we have a cycle in this graph because we can always choose to go through edges internal to $D_1$ and loop indefinitely). Therefore we can just remove that node to arrive at a new boundary (now its incoming edges are in the cut set). By the same argument, we can keep removing nodes until $D=\{s\}$, completing the proof.

\item Shapley Flow gives boundary consistent attributions: 
 We show that every boundary grown has edge attribution consistent with the previous boundary. Therefore all boundaries have consistent edge attribution because the boundary formed by any two boundary's common set of nodes can be grown into those two boundaries using the property above. Let's focus on the newly added node $c$ from one boundary to the next. Note that a property of depth first search is that every time $c$'s value is updated, its outgoing edges are activated in an atomic way (no other activation of edges occur between the activation of $c$'s outgoing edges). Therefore, the change in output due to the activation of new edges occur together in the view of edges upstream of $c$, thus not changing their attributions. Also, since $c$'s outgoing edges must point to the model defined by the current boundary (otherwise it cannot be a valid topological ordering), they don't have down stream edges, concluding the proof.

\item $\Pi^{\text{dfs}}_{\mathcal{B}}=\tilde{\mathcal{H}}_{\mathcal{B}}$: Since attribution is boundary consistent, we can treat the model as a blackbox and only look at the DFS ordering on the data side. Observe that the edge traversal ordering in DFS is a valid history because a) every edge traversal can be understood as a message received through edge , b) when every message is received, the node's value is updated, and c) the new node's value is sent out through every outgoing edge by the recursive call in DFS. Therefore the two side of the equation are at least holding the same type of object. 

We first show that $\Pi^{\text{dfs}}_{\mathcal{B}} \subseteq \tilde{\mathcal{H}}_{\mathcal{B}}$. Take $h \in \Pi^{\text{dfs}}_{\mathcal{B}}$, we need to find a history $h^*$ in $\mathcal{B}^*$ such that a) $h$ can be expanded into $h^*$ and b) for any boundary, there is a history in that boundary that can be expanded into $h^*$. Let $h^*$ be any history expanded using DFS that is aligned with $h$. To show that every boundary can expand into $h^*$, we just need to show that the boundaries generated through the growing process introduced in the first bullet point can be expanded into $h^*$. The base case is $D=\{s\}$. There must have an ordering to expand into $h^*$ because $h^*$ is generated by DFS, and that DFS ensures that every edge's impact on the boundary is propagated to the end of computation before another edge in $D$ is traversed. Similarly, for the inductive step, when a new node $c$ is added, we just follow the expansion of its previous boundary to reach $h^*$.

Next we show that $\tilde{\mathcal{H}}_{\mathcal{B}} \subseteq \Pi^{\text{dfs}}_{\mathcal{B}}$. First observe that for history $h_1$ in $\mathcal{B}_1=(D_1, F_1)$ and history $h_2$ in $\mathcal{B}_2=(D_2, F_2)$ with $F_2 \subseteq F_1$, if $h_1$ cannot be expanded into $h_2$, then $HE(h_1) \cap HE(h_2) = \emptyset$ because they already have mismatches for histories that doesn't involve passing through $\mathcal{B}_1$. Assume we do have $h \in \tilde{\mathcal{H}}_{\mathcal{B}}$ but $h \not \in \Pi^{\text{dfs}}_{\mathcal{B}}$. To derive a contradiction, we shrink the boundary one node at a time from $\mathcal{B}$, again using the procedure described in the first bullet point. We denote the resulting boundary formed by removing $n$ nodes as $\mathcal{B}_{-n}$. Since $h$ is assumed to be boundary consistent, there exist $h_{\mathcal{B}_{-1}} \in \mathcal{H}_{\mathcal{B}_{-1}}$ such that $h_{\mathcal{B}_{-1}}$ must be able to expand into $h$. Say the two boundaries differ in node $c$. Note that any update to $c$ crosses $\mathcal{B}_{-1}$, therefore its impact must be reached by $F$ before another event occurs in $D_{-1}$. Since all of $c$'s outgoing edges crosses $\mathcal{B}$, any ordering of messages sent through those edges is a DFS ordering from $c$. This means that if $h_{\mathcal{B}_{-1}}$ can be reached by DFS, so can $h_{\mathcal{B}}$, violating the assumption. Therefore, $h_{\mathcal{B}_{-1}} \not \in \Pi^{\text{dfs}}_{\mathcal{B}_{-1}}$ and $h_{\mathcal{B}_{-1}} \in \tilde{\mathcal{H}}_{\mathcal{B}_{-1}}$ (the latter because $h_{\mathcal{B}_{-1}}$ can expand into a history that is consistent with all boundaries by first expanding into $h$). We run the same argument until $D=\{s\}$. This gives a contradiction because in this boundary, all histories can be produced by DFS.

\item Efficiency: Since we are attributing credit by the change in the target node's value following a history $h$ given by DFS, the target for this particular DFS run is thus $\nu_{\mathcal{B}}(h) - \nu_{\mathcal{B}}([])$. Average over all DFS runs and noting that $\tilde{\mathcal{H}}_{\mathcal{B}} = \Pi^{\text{dfs}}_{\mathcal{B}}$ gives the target $\sum_{h \in \tilde{\mathcal{H}}_{\mathcal{B}}} \nu_{\mathcal{B}}(h) / |\tilde{\mathcal{H}}_{\mathcal{B}}| - \nu_{\mathcal{B}}([])$. Noting that each update in the target node's value must flow through one of the boundary edges. Therefore the sum of boundary edges' attribution equals to the target.
\item Linearity: For two games of the same boundary $v$ and $u$, following any history, the sum of output differences between the two games is the output difference of the sum of the two games, therefore $\phi_{v+u}$ would not differ from $\phi_v + \phi_u$. It's easy to see that extending addition to any linear combination wouldn't matter.
\item Dummy player: Since Shapley Flow is boundary consistent, we can just run DFS up to the boundary (treat $F$ as a blackbox). Since every step in DFS remains in the coalition $\tilde{\mathcal{C}}_{\mathcal{B}}$ because $\Pi^{\text{dfs}}_{\mathcal{B}} \subseteq \tilde{\mathcal{H}}_{\mathcal{B}}$, if an edge is dummy, every time it is traversed through by DFS, the output won't change by definition, thus giving it $0$ credit.
\end{itemize}

\end{proof}

Therefore Shapley Flow uniquely satisfies the axioms. We note that efficiency requirement simplifies to $f(\boldsymbol{x}) - f(\boldsymbol{x'})$ when applying it to an actual model because all histories from DFS would lead the target node to its target value. We can prove a stronger claim that actually all nodes would reach its target value when DFS finishes. To see that, we do an induction on a topological ordering of the nodes. The source nodes reaches its final value by definition. Assume this holds for the $k^{\text{th}}$ node. For the $k+1^{\text{th}}$ node, its parents achieves target value by induction. Therefore DFS would make the parents' final values visible to this node, thus updating it to the target value.

\section{Causal graphs}

While the nutrition dataset is introduced in the main text, we describe an additional dataset to further demonstrate the usefulness of Shapley Flow. Moreover, we describe in detail how the causal relationship is estimated. The resulting causal graphs for the nutrition dataset and the income dataset are visualized in \textbf{Figure \ref{fig:causal_graph}}.

\subsection{The Census Income dataset}

\begin{figure}%[H]
\centering
\subfloat[Causal graph for the nutrition dataset]{\label{fig:nutrition}
\includegraphics[width=1\linewidth,height=0.2\textheight]
{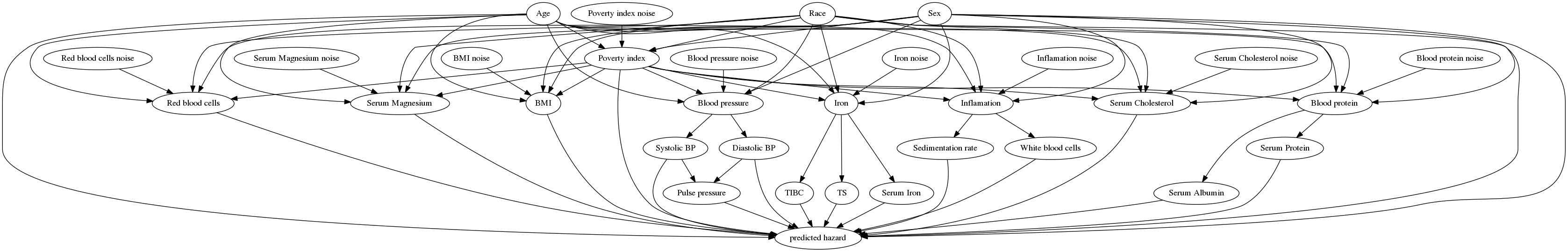}}

\subfloat[Causal graph for the Census Income dataset]{\label{fig:income}
\includegraphics[width=0.9\linewidth]{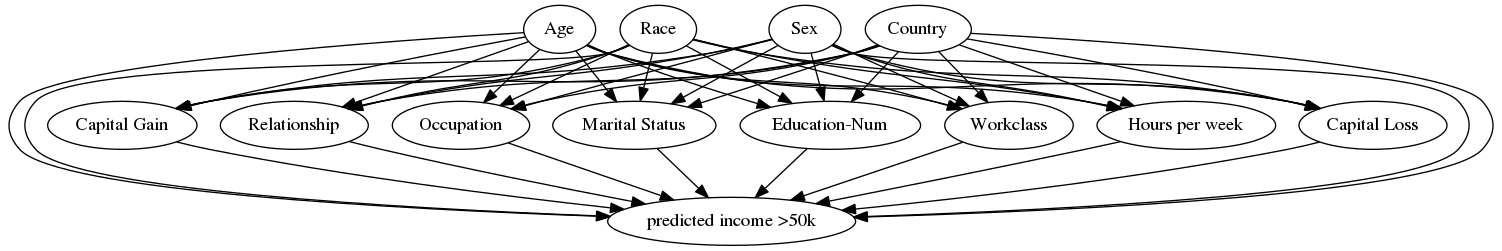}}
\caption{The causal graphs we used for the two real datasets. Note that each node in the causal graph for (a) is given a noise node to account for random effects. The noise nodes are omitted for better readability for (b). The  resulting  causal  structures  are  over-simplifications of the true causal structure; the relationship between source nodes (e.g.,  race and sex) and other features is far more complex. They are used as a proof of concept to show both the direct and indirect effect of features on the prediction output.}
\label{fig:causal_graph}
\end{figure}

The Census Income dataset consists of $32,561$ samples with $12$ features. The task is to predict whether one's annual income exceeds $50k$. We assume a causal graph, similar to that used by \cite{frye2019asymmetric} (\textbf{Figure \ref{fig:income}}). Attributes determined at birth e.g.,  sex, native country, and race act as source nodes. The remaining features (marital status, education, relationship, occupation, capital gain, work hours per week, capital loss, work class) have fully connected edges pointing from their causal ancestors. All features have a directed edge pointing to the model. 

\subsection{Causal Effect Estimation}
\label{sec:causal_detail}

Given the causal structure described above, we estimate the relationship among variables using XGBoost. More specifically, using an $80/20$ train test split, we use XGBoost to learn the function for each node.  If the node value is categorical, we train to minimize cross entropy loss. Otherwise, we minimize mean squared error. Models are fitted by $100$ XGBoost trees with a max depth of $3$ for up to $1000$ epochs. Since features are rarely perfectly determined by their dependency node, we add independent noise nodes to account for this effect. That is, each non-sink node is pointed to by a unique noise node that account for the residue effect of the prediction. % To make sure that $i$ achieves its background (foreground) value when all of its input are at their background (foreground) value, for a noise node attached to a non-categorical variable $i$, its background (foreground) value is set to be the difference between $i$'s background (foreground) value, and the $i$'s function computed with the background (foreground) value of $i$'s dependent variables. A noise node associated with a categorical feature $i$ is similarly set.

Depending on whether the variable is discrete or continuous, we handle the noise differently. For continuous variables, the noise node's value is the residue between the prediction and the actual value. For discrete variables, we assume the actual value is sampled from the categorical distribution specified by the prediction. Therefore the noise node's value is any possible random number that could result in the actual value. As a concrete example for handling discrete variable, consider a binary variable $y$, and assume the trained categorical function $f$ gives $f(\boldsymbol{x}) = [0.3, 0.7]$
where $\boldsymbol{x}$ is the foreground value of the input to predict $y$. We view the data generation as the following. The noise term associated with $y$ is treated as a uniform random variable between $0$ and $1$. If it lands within $0$ to $0.3$, $y$ is sampled to be $0$, otherwise $1$ (matching the categorical function of $70\%$ chance of sampling $y$ to be $1$). Now if we observe the foreground value of $y$ to be $0$, it means the foreground value of noise must be uniform between $0$ to $0.3$. Although we cannot infer the exact value of the noise, we can sample the noise from $0$ to $0.3$ multiple times and average the resulting attribution.

\section{Additional Results}

In this section, we first present additional sanity checks with synthetic data. Then we show additional examples from both the nutrition and income datasets to demonstrate how a complete view of boundaries should be preferable over single boundary approaches.

\subsection{Additional Sanity Checks}

\begin{figure}
    \centering

    \subfloat[chain dataset]{\label{fig:chain_graph}
    \includegraphics[width=0.3 \linewidth]{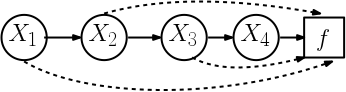}}

\begin{tabular}{lrrr}
\toprule
{} & Independent & On-manifold & ASV\\
\midrule
X4 & \cellcolor{blue!100} -1.82 & \cellcolor{blue!100} -0.45 & \cellcolor{red!0} 0.0\\
X1 & \cellcolor{red!0} 0.0 & \cellcolor{blue!100} -0.45 & \cellcolor{blue!100} -1.82\\
X3 & \cellcolor{red!0} 0.0 & \cellcolor{blue!100} -0.45 & \cellcolor{red!0} 0.0\\
X2 & \cellcolor{red!0} 0.0 & \cellcolor{blue!100} -0.45 & \cellcolor{red!0} 0.0\\
\bottomrule
\end{tabular}
  
    \subfloat[Shapley Flow]{\label{fig:chain_flow}
    \includegraphics[width=0.5\linewidth]{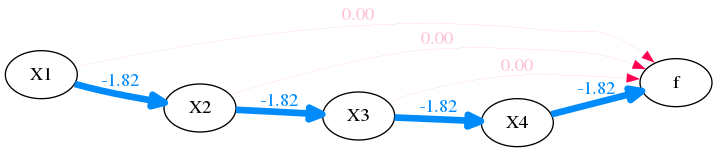}}
    
    \caption{\textbf{(a)} The chain dataset contains exact copies of nodes. The dashed edges denotes dummy dependencies. \textbf{(b)} While Shapley Flow shows the entire path of influence, other baselines fails to capture either direct and indirect effects.}
\label{fig:chain_viz}
\end{figure}

We include further sanity check experiments in this section. The first sanity check consists of a chain with 4 variables. Each node along the chain is an identical copy of its predecessor and the function to explain only depends on $X_4$ (\textbf{Figure \ref{fig:chain_viz}}). The dataset is created by sampling $X_1 \sim \mathcal{N}(0, 1)$, that is a standard normal distribution, with $1000$ samples. We use the first sample as background, and explain the second sample (one can choose arbitrary samples to obtain the same insights). As shown in \textbf{Figure \ref{fig:chain_viz}}, independent SHAP fails to show the indirect impact of $X_1$, $X_2$, and $X_3$, ASV fails to show the direct impact of $X_4$, on manifold SHAP fails to fully capture both the direct and indirect importance of any edge.

\begin{table}
\centering
\scalebox{0.8}{
\begin{tabular}{lrrr}
\toprule
{Methods} &   Income & Nutrition  &  Synthetic \\
\midrule
Independent  &  {\bf 0.0} ($\pm$ 0.0) & {\bf 0.0} ($\pm$ 0.0) & {\bf0.0} ($\pm$ 0.0) \\
On-manifold & 0.4 ($\pm$ 0.3) &             1.3 ($\pm$ 2.5) &   0.8 ($\pm$ 0.7)\\
ASV & 0.4 ($\pm$ 0.6) & 1.5 ($\pm$ 3.3) & 1.2 ($\pm$ 1.4) \\
Shapley Flow & {\bf 0.0} ($\pm$ 0.0) & {\bf 0.0} ($\pm$ 0.0) & {\bf0.0} ($\pm$ 0.0)\\
\bottomrule
\end{tabular}
}
\caption{Shapley Flow and independent SHAP have lower mean absolute error (std) for direct effect of features on linear models. %For comparison, the standard deviation in output is 1.6 for the income dataset, 2.7 for the nutrition dataset, and 3.1 for the synthetic dataset
}
\label{tab:direct_effect}
\end{table}

\begin{table}
\centering
\scalebox{0.8}{
\begin{tabular}{lrrr}
\toprule
{Methods} &   Income & Nutrition  &  Synthetic \\
\midrule
Independent  &  0.1 ($\pm$ 0.2) & 0.8 ($\pm$ 2.7) & 1.1 ($\pm$ 1.4) \\
On-manifold & 0.4 ($\pm$ 0.3) &             0.9 ($\pm$ 1.6) &   1.5 ($\pm$ 1.5)\\
ASV & 0.1 ($\pm$ 0.1) & 0.6 ($\pm$ 1.9) & 1.1 ($\pm$ 1.5) \\
Flow & {\bf 0.0} ($\pm$ 0.0) & {\bf 0.0} ($\pm$ 0.0) & {\bf0.0} ($\pm$ 0.0)\\
\bottomrule
\end{tabular}
}
\caption{Shapley Flow and ASV have lower mean absolute error (std) for indirect effect on linear models. %For comparison, the standard deviation in output is 1.6 for the income dataset, 2.7 for the nutrition dataset, and 3.1 for the synthetic dataset
}
\label{tab:indirect_effect}
\end{table}

The second sanity check consists of linear models as described in \textbf{Section \ref{sec:sanity_linear}}. We include the full result with the income dataset added in \textbf{Table \ref{tab:direct_effect}} and \textbf{Table \ref{tab:indirect_effect}} for direct and indirect effects respectively. The trend for the income dataset algins with the nutrition and synthetic dataset: only Shapley Flow makes no mistake for estimating both direct and indirect impact. Independent Shap only does well for direct effect. ASV only does well for indirect effects (it only reaches zero error when evaluated on source nodes).

\subsection{Additional examples}

\begin{figure}
    \centering
    \scalebox{0.8}{
\begin{tabular}{lrrr}
\toprule
{Top features} &   Age &  Serum Magnesium &  Race \\
\midrule
Background sample  &  35.0 &             1.37 &   Black \\
Foreground sample &  42.0 &             1.63 &   white \\
\bottomrule
\end{tabular}
}

\centering

\small{
 \scalebox{0.8}{
\begin{tabular}{lrrr}
\toprule
{Attributions} & Independent & On-manifold & ASV\\
\midrule
Age & \cellcolor{red!100} 0.23 & \cellcolor{blue!100} -0.38 & \cellcolor{red!100} 0.3\\
Serum Magnesium & \cellcolor{blue!92} -0.21 & \cellcolor{blue!4} -0.02 & \cellcolor{blue!48} -0.15\\
Race & \cellcolor{blue!26} -0.06 & \cellcolor{red!11} 0.04 & \cellcolor{blue!80} -0.24\\
Pulse pressure & \cellcolor{red!0} 0.0 & \cellcolor{blue!21} -0.08 & \cellcolor{red!0} 0.0\\
Diastolic BP & \cellcolor{red!0} 0.0 & \cellcolor{red!21} 0.08 & \cellcolor{red!0} 0.0\\
Serum Cholesterol & \cellcolor{red!1} 0.0 & \cellcolor{red!17} 0.07 & \cellcolor{red!1} 0.0\\
Serum Protein & \cellcolor{red!3} 0.01 & \cellcolor{red!15} 0.06 & \cellcolor{red!0} 0.0\\
Serum Iron & \cellcolor{red!0} 0.0 & \cellcolor{red!14} 0.05 & \cellcolor{red!0} 0.0\\
Poverty index & \cellcolor{blue!10} -0.02 & \cellcolor{red!1} 0.01 & \cellcolor{blue!3} -0.01\\
Systolic BP & \cellcolor{blue!12} -0.03 & \cellcolor{blue!1} -0.01 & \cellcolor{red!0} 0.0\\
Red blood cells & \cellcolor{red!0} 0.0 & \cellcolor{red!12} 0.05 & \cellcolor{red!0} 0.0\\
Blood protein & \cellcolor{red!0} 0.0 & \cellcolor{red!0} 0.0 & \cellcolor{red!11} 0.04\\
TIBC & \cellcolor{red!0} 0.0 & \cellcolor{red!11} 0.04 & \cellcolor{red!0} 0.0\\
Blood pressure & \cellcolor{red!0} 0.0 & \cellcolor{red!0} 0.0 & \cellcolor{blue!8} -0.03\\
TS & \cellcolor{red!0} 0.0 & \cellcolor{red!7} 0.03 & \cellcolor{red!0} 0.0\\
BMI & \cellcolor{red!0} -0.0 & \cellcolor{blue!7} -0.03 & \cellcolor{red!0} -0.0\\
Sex & \cellcolor{red!0} 0.0 & \cellcolor{red!4} 0.02 & \cellcolor{red!0} 0.0\\
Serum Albumin & \cellcolor{red!0} 0.0 & \cellcolor{blue!3} -0.01 & \cellcolor{red!0} 0.0\\
White blood cells & \cellcolor{red!2} 0.01 & \cellcolor{blue!1} -0.01 & \cellcolor{red!0} 0.0\\
Sedimentation rate & \cellcolor{red!0} 0.0 & \cellcolor{red!2} 0.01 & \cellcolor{red!0} 0.0\\
Inflamation & \cellcolor{red!0} 0.0 & \cellcolor{red!0} 0.0 & \cellcolor{red!2} 0.01\\
Iron & \cellcolor{red!0} 0.0 & \cellcolor{red!0} 0.0 & \cellcolor{red!0} 0.0\\
\bottomrule
\end{tabular}
}
}

    % \subfloat[SHAP on the nutrition data]{\label{fig:nutrition_shap}
    % \includegraphics[width=0.9 \linewidth]{figs/nutrition_shap_55.png}}
    
    % \subfloat[ASV on the nutrition data]{\label{fig:nutrition_asv}
    % \includegraphics[width=0.9\linewidth]{figs/nutrition_asv_55.png}}
  
    \subfloat[Shapley Flow]{\label{fig:nutrition_flow}
    \includegraphics[width=0.6\linewidth]{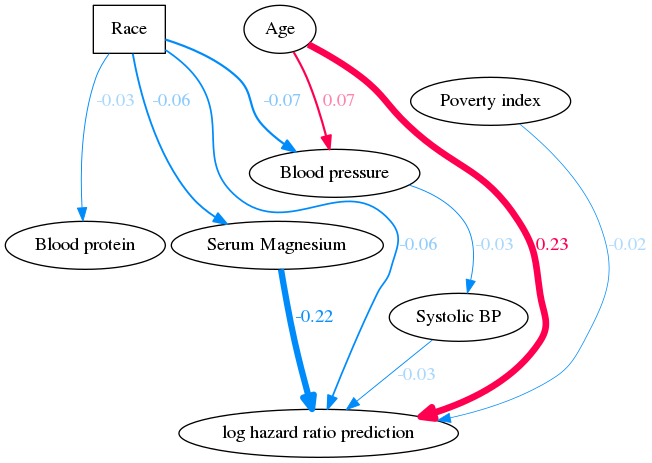}}
    
    \setlength{\belowcaptionskip}{-20pt}
    \caption{Comparison among baselines on a sample (top table) from the nutrition dataset, showing top 10 features/edges. As noted in the main text this graph is an oversimplification and is not necessarily representative of the true underlying causal relationship.}
\label{fig:nutrition_viz}
\end{figure}

In this section, we analyze another example from the nutrition dataset (\textbf{Figure \ref{fig:nutrition_viz}}) and 3 additional example from the adult censor dataset.

\textbf{Independent SHAP ignores the indirect impact of features}. Take an example from the nutrition dataset (\textbf{Figure  \ref{fig:nutrition_viz}}). The race feature is given low attribution with independent SHAP, but high importance in ASV. This happens because race, in addition to its direct impact, indirectly affects the output through blood pressure, serum magnesium, and blood protein, as shown by Shapley Flow (\textbf{Figure \ref{fig:nutrition_flow}}). In particular, race partially accounts for the impact of serum magnesium because changing race from Black to White on average increases serum magnesium by $0.07$ meg/L in the dataset (thus partially explaining the increase in serum magnesium changing from the background sample to the foreground). Independent SHAP fails to account for the indirect impact of race, leaving the user with a potentially misleading impression that race is irrelevant for the prediction. 

\begin{figure}
    \centering
    
    \subfloat[Age vs. output]{\label{fig:nutrition_age}
    \includegraphics[width=0.48 \linewidth]{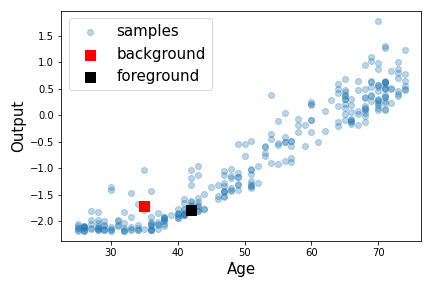}}
    \subfloat[Age vs. magnesium]{\label{fig:nutrition_age_serum_magnesium}
    \includegraphics[width=0.48\linewidth]{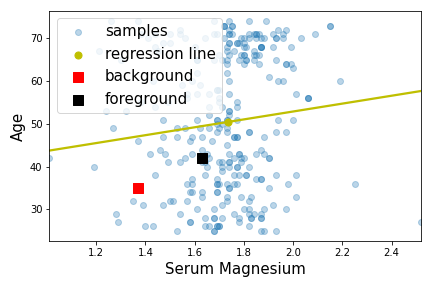}}
    % \subfloat[Shapley Flow on the nutrition data]{\label{fig:nutrition_age_serum_cholesterol}
    %\includegraphics[width=0.31\linewidth]{figs/manifold_explain_age_Serum_Cholesterol.png}}
    \setlength{\belowcaptionskip}{-10pt}

    \caption{Age appears to be protective in on-manifold SHAP because it steals credit from other variables.}
\label{fig:nutrition_explain_on_manifold}
\end{figure}

\textbf{On-manifold SHAP provides a misleading interpretation}. With the same example (\textbf{Figure  \ref{fig:nutrition_viz}}), we observe that on-manifold SHAP strongly disagrees with independent SHAP, ASV, and Shapley Flow on the importance of age. Not only does it assign more credit to age, it also flips the sign, suggesting that age is protective. However, \textbf{Figure \ref{fig:nutrition_age}} shows that age and earlier mortality are positively correlated; then how could age be protective?  \textbf{Figure \ref{fig:nutrition_age_serum_magnesium}} provides an explanation. Since SHAP considers all partial histories regardless of the causal structure, when we focus on serum magnesium and age, there are two cases: serum magnesium updates before or after age. We focus on the first case because it is where on-manifold SHAP differs from other baselines (all baselines already consider the second case as it satisfies the causal ordering). When serum magnesium updates before age, the expected age given serum magnesium is higher than the foreground age (yellow line above the black marker). Therefore when age updates to its foreground value, we observe a decrease in age, leading to a decrease in the output (so age appears to be protective). Serum magnesium is just one variable from which age steals credit. Similar logic applies to  TIBC, red blood cells, serum iron, serum protein, serum cholesterol, and diastolic BP. From both an in/direct impact perspective, on-manifold perturbation can be misleading since it is based not on causal but on observational relationships.

\textbf{ASV ignores the direct impact of features}. As shown in \textbf{Figure \ref{fig:nutrition_viz}}, serum magnesium appears to be more important in independent SHAP compared to ASV. From Shapley Flow (\textbf{Figure \ref{fig:nutrition_flow}}), this difference is explained by race as its edge to serum magnesium has a negative impact. However, looking at ASV alone, one fails to understand that intervening on serum magnesium could have a larger impact on the output.

\textbf{Shapley Flow shows both direct and indirect impacts of features}. Focusing on the attribution given by Shapley Flow (\textbf{Figure \ref{fig:nutrition_flow}}). We not only observe similar direct impacts in variables compared to independent SHAP, but also can trace those impacts to their source nodes, similar to ASV. Furthermore, Shapley Flow provides more detail compared to other approaches. For example, using Shapley Flow we gain a better understanding of the ways in which race impacts survival. The same goes for all other features. This is useful because causal links can change (or break) over time. Our method provides a way to reason through the impact of such a change.

\begin{figure}
    \centering

\small{
\begin{tabular}{lll}
\toprule
{} & Background sample &    Foreground sample \\
\midrule
Age            &                39 &                   35 \\
Workclass      &         State-gov &          Federal-gov \\
Education-Num  &                13 &                    5 \\
Marital Status &     Never-married &   Married-civ-spouse \\
Occupation     &      Adm-clerical &      Farming-fishing \\
Relationship   &     Not-in-family &              Husband \\
Race           &             White &                Black \\
Sex            &              Male &                 Male \\
Capital Gain   &              2174 &                    0 \\
Capital Loss   &                 0 &                    0 \\
Hours per week &                40 &                   40 \\
Country        &     United-States &        United-States \\
\bottomrule
\end{tabular}
}
\begin{tabular}{lrrr}
\toprule
{} & Independent & On-manifold & ASV\\
\midrule
Education-Num & \cellcolor{blue!100} -0.12 & \cellcolor{blue!100} -0.11 & \cellcolor{blue!100} -0.09\\
Relationship & \cellcolor{red!42} 0.05 & \cellcolor{red!57} 0.06 & \cellcolor{red!49} 0.04\\
Capital Gain & \cellcolor{red!81} 0.09 & \cellcolor{red!7} 0.01 & \cellcolor{red!29} 0.03\\
Occupation & \cellcolor{blue!28} -0.03 & \cellcolor{blue!68} -0.07 & \cellcolor{blue!20} -0.02\\
Marital Status & \cellcolor{red!36} 0.04 & \cellcolor{red!48} 0.05 & \cellcolor{red!28} 0.03\\
Workclass & \cellcolor{red!19} 0.02 & \cellcolor{red!32} 0.03 & \cellcolor{red!20} 0.02\\
Race & \cellcolor{blue!7} -0.01 & \cellcolor{blue!27} -0.03 & \cellcolor{red!14} 0.01\\
Age & \cellcolor{blue!10} -0.01 & \cellcolor{blue!4} -0.01 & \cellcolor{red!20} 0.02\\
Capital Loss & \cellcolor{red!0} 0.0 & \cellcolor{red!31} 0.03 & \cellcolor{red!0} 0.0\\
Country & \cellcolor{red!0} 0.0 & \cellcolor{red!28} 0.03 & \cellcolor{red!0} 0.0\\
Sex & \cellcolor{red!0} 0.0 & \cellcolor{red!26} 0.03 & \cellcolor{red!0} 0.0\\
Hours per week & \cellcolor{red!0} 0.0 & \cellcolor{red!3} 0.0 & \cellcolor{red!0} 0.0\\
\bottomrule
\end{tabular}

    \subfloat[Shapley Flow]{
    \includegraphics[width=0.9\linewidth]{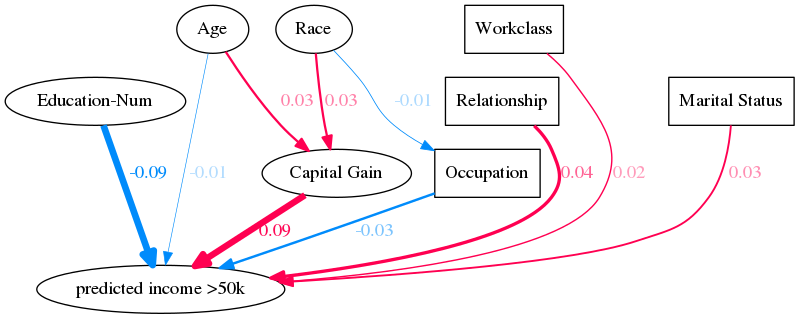}}
    \caption{Comparison between independent SHAP, on-manifold SHAP, ASV, and Shapley Flow on a sample from the income dataset. Shapley flow shows the top 10 links. The direct impact of capital gain is not represented by on-manifold SHAP. As noted in the text this graph is based on previous work and is not necessarily representative of the true underlying causal relationship.}
\label{fig:income_viz_22}
\end{figure}

\begin{figure}
    \centering

\small{
\begin{tabular}{lll}
\toprule
{} &              Background sample  &                   foreground sample \\
\midrule
Age            &              39 &                   30 \\
Workclass      &       State-gov &            State-gov \\
Education-Num  &              13 &                   13 \\
Marital Status &   Never-married &   Married-civ-spouse \\
Occupation     &    Adm-clerical &       Prof-specialty \\
Relationship   &   Not-in-family &              Husband \\
Race           &           White &   Asian-Pac-Islander \\
Sex            &            Male &                 Male \\
Capital Gain   &            2174 &                    0 \\
Capital Loss   &               0 &                    0 \\
Hours per week &              40 &                   40 \\
Country        &   United-States &                India \\
\bottomrule
\end{tabular}
}

\begin{tabular}{lrrr}
\toprule
{} & Independent & On-manifold & ASV\\
\midrule
Relationship & \cellcolor{red!76} 0.17 & \cellcolor{red!31} 0.04 & \cellcolor{red!96} 0.13\\
Capital Gain & \cellcolor{red!100} 0.22 & \cellcolor{red!6} 0.01 & \cellcolor{red!53} 0.07\\
Occupation & \cellcolor{red!44} 0.1 & \cellcolor{red!50} 0.06 & \cellcolor{red!51} 0.07\\
Marital Status & \cellcolor{red!37} 0.08 & \cellcolor{red!47} 0.06 & \cellcolor{red!51} 0.07\\
Country & \cellcolor{blue!19} -0.04 & \cellcolor{red!55} 0.07 & \cellcolor{red!53} 0.07\\
Age & \cellcolor{red!0} -0.0 & \cellcolor{blue!13} -0.02 & \cellcolor{red!100} 0.13\\
Education-Num & \cellcolor{red!0} 0.0 & \cellcolor{red!100} 0.12 & \cellcolor{red!0} 0.0\\
Race & \cellcolor{red!8} 0.02 & \cellcolor{red!60} 0.07 & \cellcolor{red!1} 0.0\\
Workclass & \cellcolor{red!0} 0.0 & \cellcolor{red!50} 0.06 & \cellcolor{red!0} 0.0\\
Hours per week & \cellcolor{red!0} 0.0 & \cellcolor{red!22} 0.03 & \cellcolor{red!0} 0.0\\
Sex & \cellcolor{red!0} 0.0 & \cellcolor{red!22} 0.03 & \cellcolor{red!0} 0.0\\
Capital Loss & \cellcolor{red!0} 0.0 & \cellcolor{red!11} 0.01 & \cellcolor{red!0} 0.0\\
\bottomrule
\end{tabular}
    \subfloat[Shapley Flow]{
    \includegraphics[width=0.9\linewidth]{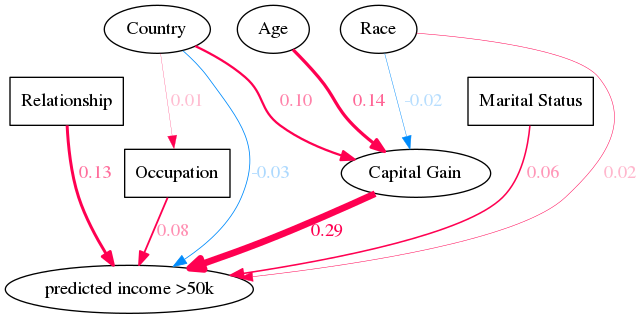}}
    \caption{Comparison between independent SHAP, on-manifold SHAP, ASV, and Shapley Flow on a sample from the income dataset. Shapley flow shows the top 10 links. The indirect impact of age is only highlighted by Shapley Flow and ASV. As noted in the text this graph is based on previous work and is not necessarily representative of the true underlying causal relationship.}
\label{fig:income_viz_11}
\end{figure}

\begin{figure}
    \centering

\small{
\begin{tabular}{lll}
\toprule
{} & Background sample &    Foreground sample \\
\midrule
Age            &                39 &                   30 \\
Workclass      &         State-gov &          Federal-gov \\
Education-Num  &                13 &                   10 \\
Marital Status &     Never-married &   Married-civ-spouse \\
Occupation     &      Adm-clerical &         Adm-clerical \\
Relationship   &     Not-in-family &            Own-child \\
Race           &             White &                White \\
Sex            &              Male &                 Male \\
Capital Gain   &              2174 &                    0 \\
Capital Loss   &                 0 &                    0 \\
Hours per week &                40 &                   40 \\
Country        &     United-States &        United-States \\
\bottomrule
\end{tabular}
}

\begin{tabular}{lrrr}
\toprule
{Attributions} & Independent & On-manifold & ASV\\
\midrule
Marital Status & \cellcolor{red!58} 0.03 & \cellcolor{red!69} 0.08 & \cellcolor{red!100} 0.03\\
Capital Gain & \cellcolor{red!100} 0.06 & \cellcolor{red!15} 0.02 & \cellcolor{red!71} 0.02\\
Workclass & \cellcolor{red!50} 0.03 & \cellcolor{red!26} 0.03 & \cellcolor{red!74} 0.02\\
Relationship & \cellcolor{blue!13} -0.01 & \cellcolor{blue!100} -0.11 & \cellcolor{red!32} 0.01\\
Education-Num & \cellcolor{blue!40} -0.02 & \cellcolor{red!11} 0.01 & \cellcolor{blue!68} -0.02\\
Age & \cellcolor{blue!38} -0.02 & \cellcolor{blue!30} -0.03 & \cellcolor{red!22} 0.01\\
Country & \cellcolor{red!0} 0.0 & \cellcolor{red!29} 0.03 & \cellcolor{red!0} 0.0\\
Capital Loss & \cellcolor{red!0} 0.0 & \cellcolor{red!25} 0.03 & \cellcolor{red!0} 0.0\\
Occupation & \cellcolor{red!0} 0.0 & \cellcolor{blue!23} -0.03 & \cellcolor{red!0} 0.0\\
Sex & \cellcolor{red!0} 0.0 & \cellcolor{red!23} 0.03 & \cellcolor{red!0} 0.0\\
Race & \cellcolor{red!0} 0.0 & \cellcolor{red!14} 0.02 & \cellcolor{red!0} 0.0\\
Hours per week & \cellcolor{red!0} 0.0 & \cellcolor{blue!2} -0.0 & \cellcolor{red!0} 0.0\\
\bottomrule
\end{tabular}

    \subfloat[Shapley Flow]{
    \includegraphics[width=0.9\linewidth]{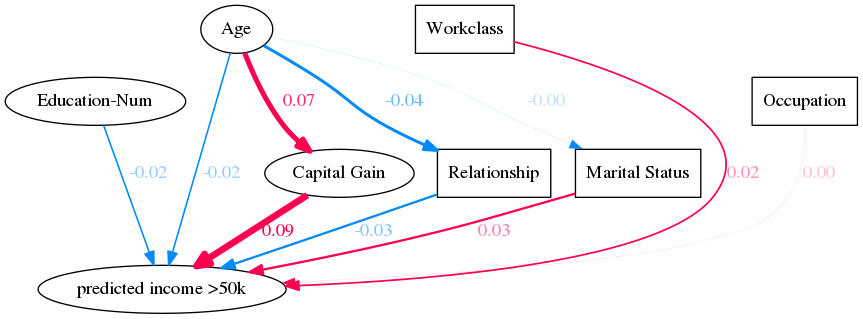}}
    \caption{Comparison between independent SHAP, on-manifold SHAP, ASV, and Shapley Flow on a sample from the income dataset. Shapley flow shows the top 10 links. Note that although age appears to be not important for all baselines, its impact through different causal edges are opposite as shown by Shapley Flow.}
\label{fig:income_viz_33}
\end{figure}

\textbf{Figure \ref{fig:income_viz_22}} gives an example of applying Shapley Flow and baselines on the income dataset. Note that the attribution to capital gain drops from independent SHAP to on-manifold SHAP and ASV. From Shapley Flow, we know the decreased attribution is due to age and race. More examples are shown in \textbf{Figure \ref{fig:income_viz_11}} and \textbf{ \ref{fig:income_viz_33}}.

%%%%%%

%%%%
%%%%%%

\subsection{A global understanding with Shapley Flow}

\begin{figure}
    \centering
    
\begin{tabular}{lrrr}
\toprule
{} & Independent & On-manifold & ASV\\
\midrule
Capital Gain & \cellcolor{red!92} 0.02 & \cellcolor{red!69} 0.02 & \cellcolor{red!100} 0.03\\
Education-Num & \cellcolor{red!100} 0.02 & \cellcolor{red!100} 0.03 & \cellcolor{red!60} 0.02\\
Age & \cellcolor{red!49} 0.01 & \cellcolor{red!45} 0.01 & \cellcolor{red!30} 0.01\\
Occupation & \cellcolor{red!21} 0.0 & \cellcolor{red!48} 0.01 & \cellcolor{red!13} 0.0\\
Capital Loss & \cellcolor{red!39} 0.01 & \cellcolor{blue!14} -0.0 & \cellcolor{red!19} 0.01\\
Relationship & \cellcolor{red!32} 0.01 & \cellcolor{red!16} 0.0 & \cellcolor{red!7} 0.0\\
Hours per week & \cellcolor{red!2} 0.0 & \cellcolor{red!44} 0.01 & \cellcolor{blue!3} -0.0\\
Sex & \cellcolor{red!16} 0.0 & \cellcolor{blue!24} -0.01 & \cellcolor{red!8} 0.0\\
Country & \cellcolor{red!2} 0.0 & \cellcolor{blue!32} -0.01 & \cellcolor{red!1} 0.0\\
Marital Status & \cellcolor{blue!10} -0.0 & \cellcolor{red!8} 0.0 & \cellcolor{blue!8} -0.0\\
Race & \cellcolor{red!3} 0.0 & \cellcolor{blue!21} -0.01 & \cellcolor{red!0} -0.0\\
Workclass & \cellcolor{red!2} 0.0 & \cellcolor{blue!8} -0.0 & \cellcolor{blue!1} -0.0\\
\bottomrule
\end{tabular}

    \subfloat[Shapley Flow]{\label{fig:income_global_flow}
    \includegraphics[width=0.9\linewidth]{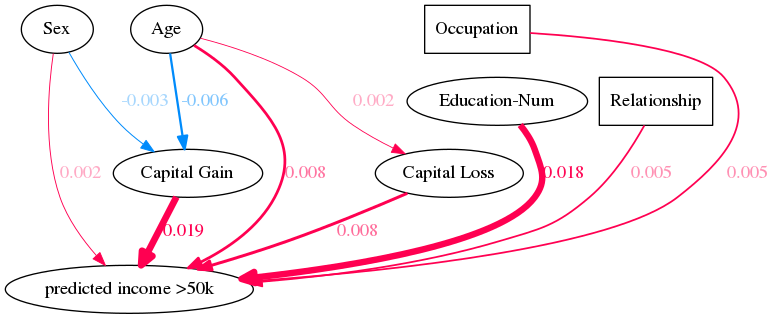}}
    
    \caption{Comparison of global understanding between independent SHAP, on-manifold SHAP, ASV, and Shapley Flow on the income dataset. Showing only the top $10$ attributions for Shapley Flow for visual clarity.}
\label{fig:income_global_viz}
\end{figure}

In addition to explaining a particular example, one can explain an entire dataset with Shapley Flow. Specifically, for multi-class classification problems, we take the average of attributions for the probability predicted for the actual class, in accordance with \citep{frye2019asymmetric}. A demonstration on the income dataset using $1000$ randomly selected examples is included in \textbf{Figure \ref{fig:income_global_viz}}. As before, we use a single shared background sample for explanation. Here, we observe that although the relative importance across independent SHAP, on-manifold SHAP, and ASV are similar, age and sex have opposite direct versus indirect impact as shown by Shapley Flow.

%However, simply average over the attributions for all examples wouldn't work because an important feature that have opposite signs of influence for different samples could cancel out, making the feature less important than it really is. One fix is to average over the absolute values of attributions. This, however, would violate the conservation of flow. To fix this problem, we change the explanation target to be the loss (assumed to be non-negative) of each prediction. 

\subsection{Example with multiple background samples}
\label{sec:multiple_background}
\begin{figure}
    \centering
    \scalebox{0.8}{
\begin{tabular}{lrrr}
\toprule
{Top features} &   Sex & Age &  Systolic BP\\
\midrule
Background mean  &  NaN & 50 &   135 \\
Foreground sample &  Female & 51 &  118 \\
\bottomrule
\end{tabular}
}

\centering

\small{
 \scalebox{0.8}{
\begin{tabular}{lrrr}
\toprule
{Attributions} & Independent & On-manifold & ASV\\
\midrule
Sex & \cellcolor{blue!100} -0.11 & \cellcolor{blue!69} -0.16 & \cellcolor{blue!100} -0.1\\
Age & \cellcolor{blue!66} -0.07 & \cellcolor{red!100} 0.23 & \cellcolor{blue!75} -0.08\\
Systolic BP & \cellcolor{blue!46} -0.05 & \cellcolor{blue!96} -0.22 & \cellcolor{red!0} 0.0\\
Poverty index & \cellcolor{blue!28} -0.03 & \cellcolor{red!37} 0.09 & \cellcolor{blue!22} -0.02\\
Blood pressure & \cellcolor{red!0} 0.0 & \cellcolor{red!0} 0.0 & \cellcolor{blue!73} -0.08\\
TIBC & \cellcolor{red!2} 0.0 & \cellcolor{blue!70} -0.16 & \cellcolor{red!0} 0.0\\
Diastolic BP & \cellcolor{blue!20} -0.02 & \cellcolor{blue!36} -0.08 & \cellcolor{red!0} 0.0\\
Pulse pressure & \cellcolor{blue!9} -0.01 & \cellcolor{blue!45} -0.11 & \cellcolor{red!0} 0.0\\
Serum Iron & \cellcolor{red!8} 0.01 & \cellcolor{red!29} 0.07 & \cellcolor{red!0} 0.0\\
BMI & \cellcolor{blue!4} -0.0 & \cellcolor{blue!21} -0.05 & \cellcolor{blue!4} -0.0\\
White blood cells & \cellcolor{blue!11} -0.01 & \cellcolor{red!14} 0.03 & \cellcolor{red!0} 0.0\\
Serum Protein & \cellcolor{blue!3} -0.0 & \cellcolor{red!21} 0.05 & \cellcolor{red!0} 0.0\\
Serum Albumin & \cellcolor{blue!3} -0.0 & \cellcolor{blue!18} -0.04 & \cellcolor{red!0} 0.0\\
Inflamation & \cellcolor{red!0} 0.0 & \cellcolor{red!0} 0.0 & \cellcolor{blue!21} -0.02\\
Serum Cholesterol & \cellcolor{blue!1} -0.0 & \cellcolor{red!15} 0.04 & \cellcolor{blue!3} -0.0\\
Iron & \cellcolor{red!0} 0.0 & \cellcolor{red!0} 0.0 & \cellcolor{red!14} 0.02\\
Sedimentation rate & \cellcolor{blue!7} -0.01 & \cellcolor{blue!3} -0.01 & \cellcolor{red!0} 0.0\\
Race & \cellcolor{blue!2} -0.0 & \cellcolor{red!1} 0.0 & \cellcolor{blue!7} -0.01\\
TS & \cellcolor{red!5} 0.01 & \cellcolor{red!5} 0.01 & \cellcolor{red!0} 0.0\\
Serum Magnesium & \cellcolor{blue!2} -0.0 & \cellcolor{blue!3} -0.01 & \cellcolor{blue!2} -0.0\\
Blood protein & \cellcolor{red!0} 0.0 & \cellcolor{red!0} 0.0 & \cellcolor{blue!6} -0.01\\
Red blood cells & \cellcolor{red!0} -0.0 & \cellcolor{red!3} 0.01 & \cellcolor{red!0} -0.0\\
\bottomrule
\end{tabular}
}
}

    \subfloat[Shapley Flow]{\label{fig:nutrition_flow_99}
    \includegraphics[width=0.6\linewidth]{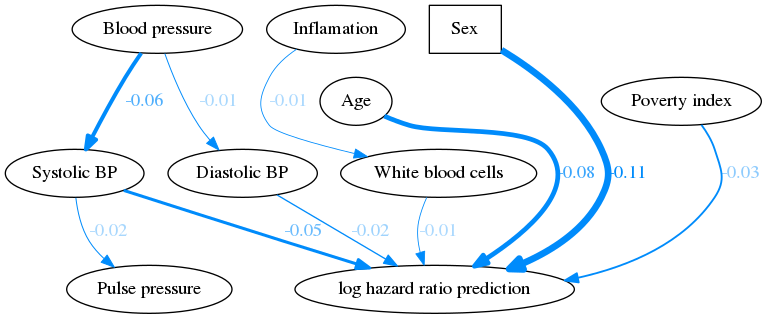}}
    
    \setlength{\belowcaptionskip}{-20pt}
    \caption{Comparison among methods on 100 background samples from the nutrition dataset, showing top 10 features/edges.}
\label{fig:nutrition_viz_99}

\end{figure}

An example with 100 background samples is shown in \textbf{Figure  \ref{fig:nutrition_viz_99}}. Shapley Flow shows a holistic picture of feature importance, while other baselines only show part of the picture.

\textbf{Independent SHAP ignores the indirect impact of features}. Take an example from the nutrition dataset (\textbf{Figure  \ref{fig:nutrition_viz_99}}). Independent SHAP only considers the direct impact of systolic blood pressure, and ignores its potential impact on pulse pressure (as shown by Shapley Flow in \textbf{Figure \ref{fig:nutrition_flow_99}}). If the causal graph is correct, independent SHAP would underestimate the effect of intervening on Systolic BP.

\begin{figure}
    \centering
    
    \subfloat[Age vs. output]{\label{fig:nutrition_age_99}
    \includegraphics[width=0.34 \linewidth]{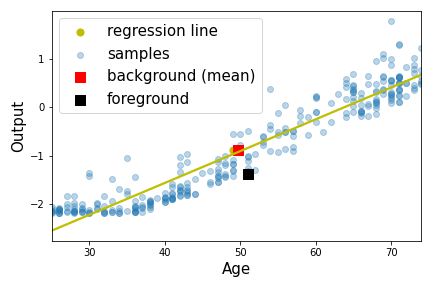}}
    \subfloat[Age vs. systolic blood pressure]{\label{fig:nutrition_age_systolic_bp_99}
    \includegraphics[width=0.34\linewidth]{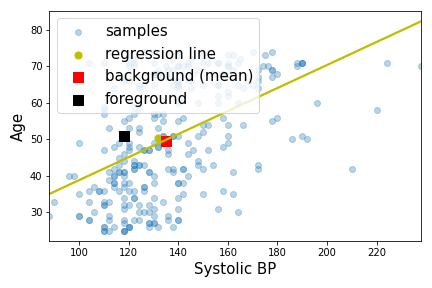}}
    \setlength{\belowcaptionskip}{-10pt}

    \caption{Age appears to be highly risky in on-manifold SHAP because it steals credit from other variables.}
\label{fig:nutrition_explain_on_manifold_99}
\end{figure}

\textbf{On-manifold SHAP provides a misleading interpretation}. With the same example (\textbf{Figure  \ref{fig:nutrition_viz_99}}), we observe that on-manifold SHAP strongly disagrees with independent SHAP, ASV, and Shapley Flow on the importance of age. In particular, it flips the sign on the importance of age. Since the background age (50) is very close to the foreground age (51), we would not expected age to significantly affect the prediction. \textbf{Figure \ref{fig:nutrition_age_systolic_bp_99}} provides an explanation. Since SHAP considers all partial histories regardless of the causal structure, when we focus on systolic blood pressure and age, there are two cases: systolic blood pressure updates before or after age. We focus on the first case because it is where on-manifold SHAP differs from other baselines (all baselines already consider the second case as it satisfies the causal ordering). When systolic blood pressure updates before age, the expected age given systolic blood pressure is lower than the foreground age (yellow line below the black marker). Therefore when age updates to its foreground value, we observe a large increase in age, leading to a increase in the output (so age appears to be riskier). from both an in/direct impact perspective, on-manifold perturbation can be misleading since it is based not on causal but on observational relationships.

\textbf{ASV ignores the direct impact of features}. As shown in \textbf{Figure \ref{fig:nutrition_viz_99}}, ASV gives no credit systolic blood pressure because it is an intermediate node. However, it is clear from Shapley Flow that intervening on systolic blood pressure has a large impact on the outcome. 

\textbf{Shapley Flow shows both direct and indirect impacts of features}. Focusing on the attribution given by Shapley Flow (\textbf{Figure \ref{fig:nutrition_flow_99}}). We not only observe similar direct impacts in variables compared to independent SHAP, but also can trace those impacts to their source nodes, similar to ASV. 

\section{Considering all histories could lead to boundary inconsistency}
\label{sec:inconsistent_history}

\begin{figure}
    \centering
    
    \subfloat[Red cut]{\label{fig:inconsistent_history_red}
    \includegraphics[width=0.48 \linewidth]{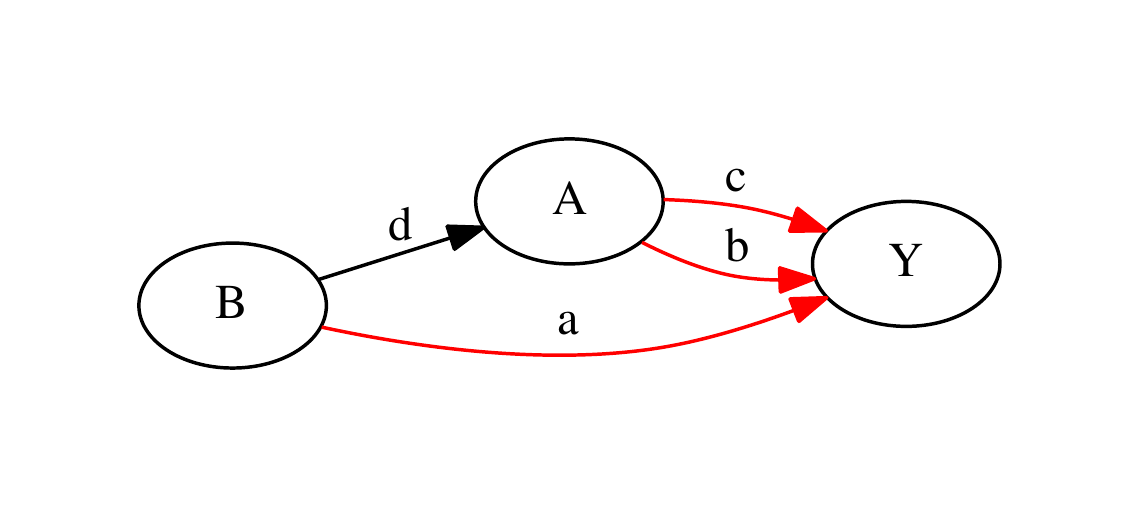}}
    \subfloat[Green cut]{\label{fig:inconsistent_hisotry_green}
    \includegraphics[width=0.48\linewidth]{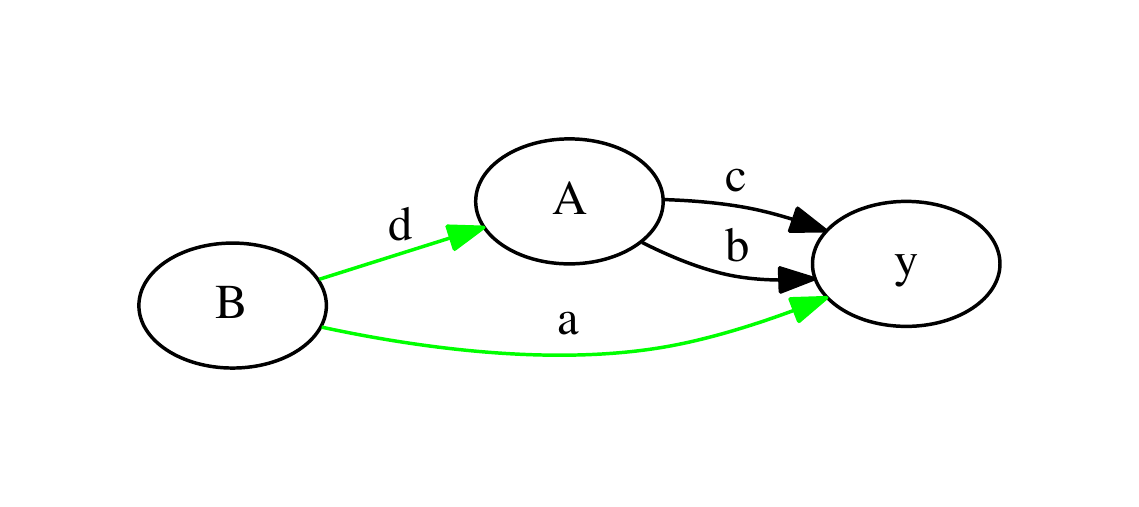}}
    \setlength{\belowcaptionskip}{-10pt}

    \caption{Two cuts that represent two boundaries for the same causal graph.}
\label{fig:inconsistent_history_example}
\end{figure}

In this section, we give an example of how considering all history $\mathcal{H}$ in the axioms (as opposed to $\mathcal{\tilde{H}}$) could lead to inconsistent attributions across boundaries. Consider two cuts for the same causal graph shown in \textbf{Figure \ref{fig:inconsistent_history_example}}. Note that both the green and the red cut share the edge ``a". We have $8$ possible message transmission histories (`c', `b' can be transmitted only after `d' has been transmitted): $\{[a, d, c, b], [a, d, b, c], [d, a, c, b], [d, a, b, c], [d, c, a, b], [d, c, b, a], [d, b, a, c], [d, b, c, a]\}$. We use the same notation for carrier games (defined in \textbf{Section \ref{sec:proof}}) and construct a game as the following:

$v_{red} = v_{red}^{dca} - v_{red}^{dcab} + v_{red}^{dba} - v_{red}^{dbac}$

Because of the linearity axiom, we have $\phi_{v_{red}}(a) > 0, \phi_{v_{red}}(b) < 0, \phi_{v_{red}}(c) < 0, \phi_{v_{red}}(d) = 0$.

However, when we consider the green boundary, the ordering $dcab$ and $dbac$ does not exist because in the green boundary $A$ and $Y$ are assumed to be a black-box. Therefore,
$v_{green} = \boldsymbol{0}$, which means $a$ is now a dummy edge: $\phi_{v_{green}}(a) = 0 \neq \phi_{v_{red}}(a)$. This demonstrate that we cannot consider all histories in $\mathcal{H}$ and being boundary consistent.

\end{document}